# Predicting the Performance of IDA* using Conditional Distributions

**Uzi Zahavi**                                        ZAHAVIU@CS.BIU.AC.IL
*Computer Science Department*
*Bar-Ilan University, Israel*

**Ariel Felner**                                        FELNER@BGU.AC.IL
*Department of Information Systems Engineering*
*Deutsche Telekom Labs*
*Ben-Gurion University, Israel*

**Neil Burch**                                        BURCH@CS.UALBERTA.CA
*Computing Science Department*
*University of Alberta, Canada*

**Robert C. Holte**                                        HOLTE@CS.UALBERTA.CA
*Computing Science Department*
*University of Alberta, Canada*

## Abstract

Korf, Reid, and Edelkamp introduced a formula to predict the number of nodes IDA* will expand on a single iteration for a given consistent heuristic, and experimentally demonstrated that it could make very accurate predictions. In this paper we show that, in addition to requiring the heuristic to be consistent, their formula's predictions are accurate only at levels of the brute-force search tree where the heuristic values obey the *unconditional* distribution that they defined and then used in their formula. We then propose a new formula that works well without these requirements, *i.e.*, it can make accurate predictions of IDA*'s performance for inconsistent heuristics and if the heuristic values in any level do not obey the unconditional distribution. In order to achieve this we introduce the *conditional distribution* of heuristic values which is a generalization of their unconditional heuristic distribution. We also provide extensions of our formula that handle individual start states and the augmentation of IDA* with bidirectional pathmax (BPMX), a technique for propagating heuristic values when inconsistent heuristics are used. Experimental results demonstrate the accuracy of our new method and all its variations.

## 1. Introduction and Overview

Heuristic search algorithms such as A* (Hart, Nilsson, & Raphael, 1968) and IDA* (Korf, 1985) are guided by the cost function $f(n) = g(n) + h(n)$, where $g(n)$ is the actual distance from the start state to state $n$ and $h(n)$ is a heuristic function estimating the cost from $n$ to the nearest goal state. A heuristic $h$ is *admissible* if $h(n) \leq dist(n, goal)$ for every state $n$ and goal state $goal$, where $dist(n, m)$ is the cost of a least-cost path from $n$ to $m$. If $h(n)$ is admissible, *i.e.* always returns a lower bound estimate of the optimal cost, these algorithms are guaranteed to find an optimal path from the start state to a goal state if one exists.

An important question to ask is how many nodes will be expanded by these algorithms to solve a given problem. A major advance in answering this question was the work done by Korf, Reid, and Edelkamp which introduced a formula to predict the number of nodes IDA* will expand (Korf & Reid, 1998; Korf, Reid, & Edelkamp, 2001). These papers, the formula they present, and the predictions it makes, will all be referred to as `KRE` in this paper. Prior to `KRE`, the standard method for comparing two heuristic functions was to compare their average values, with preference being given to the heuristic with the larger average (Korf, 1997; Korf & Felner, 2002; Felner, Korf, Meshulam, & Holte, 2007). `KRE` made a substantial improvement on this by characterizing the quality of a heuristic function by the distribution of its values. They then developed the `KRE` formula based on the heuristic distribution to predict the number of nodes expanded by IDA* when it is searching with a specific heuristic and cost threshold. Finally, they compared the predictions of their formula to the actual number of nodes expanded by IDA* for different thresholds on several benchmark search spaces and showed that it gave virtually perfect predictions. This was a major advance in the analysis of search algorithms and heuristics.

Despite its impressive results, the `KRE` formula has two main shortcomings. The first is that `KRE` assumes that in addition to being admissible the given heuristic is also *consistent*. A heuristic $h$ is *consistent* if for every pair of states, $m$ and $n$, $h(m) - h(n) \leq dist(m, n)$.[1] When the heuristic is consistent, the heuristic values of a node's children are thus constrained to be similar to the heuristic value of the node. A heuristic is *inconsistent* if it is not consistent, *i.e.* if for some pair of nodes $m$ and $n$, $h(m) - h(n) > dist(m, n)$. Inconsistency allows a node's children to have heuristic values that are arbitrarily larger or smaller than the node's own heuristic value. While the term *inconsistency* has a negative connotation as something to be avoided, recent studies have shown that inconsistent heuristics are easy to define in many search applications and can produce substantial performance improvements (Felner, Zahavi, Schaeffer, & Holte, 2005; Zahavi, Felner, Schaeffer, & Sturtevant, 2007; Zahavi, Felner, Holte, & Schaeffer, 2008). For this reason, it is important to extend the `KRE` formula to accurately predict IDA*'s performance on inconsistent heuristics, as such heuristics are likely to become increasingly important in future applications.

The second shortcoming of the `KRE` formula is that it works well only at levels of the search tree where the heuristic distribution follows the *equilibrium distribution* (defined below in Section 3.1.2). This always holds at sufficiently deep levels of the search tree, where the heuristic values converge to the *equilibrium distribution*. In addition, it will hold at **all** levels when the heuristic values of the set of start states is distributed according to the equilibrium distribution. However, as will be shown below (in Section 3.2.2) the `KRE` formula can be very inaccurate at depths of practical interest on single start states and on large sets of start states whose values are not distributed according to the equilibrium distribution. In such cases, the heuristic values at the levels of the search tree that are actually examined by IDA* will not obey the equilibrium distribution and applying `KRE` to these cases will result in inaccurate predictions.

The main objective of this paper is to develop a formula to accurately predict the number of nodes IDA* will expand, for a given cost threshold, for any given heuristic and set of start states, including those not currently covered by `KRE`. To do this we first extend `KRE`'s idea of a heuristic distribution, which is *unconditional*, to a *conditional distribution*, in which the probability of a specific heuristic value is not constant, as in `KRE`, but is conditioned on certain local properties of the search space. Our conditional distribution provides more insights about the behavior of the heuristic values during search because it is informed about when (in what context in the search tree) a specific heuristic value will be produced. This allows for a better study of heuristic behavior.

---

1. This is a general definition for any graph. In the case of undirected graphs we can write the consistency definition as $|h(m) - h(n)| \leq dist(m, n)$.

Based on the conditional distribution we develop a new formula, `CDP` (Conditional Distribution Prediction), that predicts IDA*'s performance on any set of start states (regardless of how their heuristic values are distributed) and for any desired depth (not necessarily large) whether the heuristic is consistent or not. `CDP` has a recursive structure and information about the number of nodes is propagated from the root to the leaves of the search tree. In all of our experiments `CDP`'s predictions are at least as accurate as `KRE`'s, and `CDP` is much more accurate for inconsistent heuristics or sets of start states that have non-equilibrium heuristic distributions. In its basic form, `CDP` is not particularly accurate on single start states. We describe a simple extension that improves its accuracy in this setting. Finally, we adapt `CDP` to make predictions when IDA* is augmented with the *bidirectional pathmax* method (BPMX) (Felner et al., 2005). When inconsistent heuristics are being used, BPMX is a useful addition to IDA*. It prunes many subtrees that would otherwise be explored, thereby substantially reducing the number of nodes IDA* expands.

Throughout the paper we provide experimental results demonstrating the accuracy of `CDP` in all of the above scenarios using the same two benchmark domains used in `KRE` – the sliding-tile puzzle and Rubik's Cube.

For simplicity of discussion, we assume in this paper that all edges cost 1. This is true for many problem domains. The generalization of the ideas to the case of variable edge costs is straightforward, although their practical implementation introduces some additional challenges (briefly described in Section 11.2).

The paper is organized as follows. Section 2 presents background material. Section 3 derives the `KRE` formula from first principles and discusses its limitations. In Section 4, our notion of the *conditional distribution* of heuristic values is presented. Our new formula, `CDP`, is presented in Section 4.2. Section 5 discusses a subtle but important way in which our experiments differ from `KRE`'s. Experimental results are presented in Sections 6 and 7. The extension of the `CDP` formula to better handle single start states is presented in Section 8. Section 9 proposes a technique, based on `CDP`, for estimating upper and lower bounds on the number of nodes IDA* can expand for a given unconditional distribution. Section 10 presents an extension of `CDP` for predicting the performance of IDA* when BPMX is applied. Related work is discussed in Section 11, and conclusions and suggestions for future work are given in Section 12. A preliminary version of this paper appeared (Zahavi, Felner, Burch, & Holte, 2008).

## 2. Background

Two application domains were used by `KRE` to demonstrate the accuracy of their formula. In our experiments we use exactly the same domains. In this section we describe them as well as the search algorithm and the different heuristic functions that are used in our experiments.

## 2.1 Problem Domains

Two of the classic examples in the AI literature of a single-agent pathfinding problems are Rubik's Cube and the sliding-tile puzzle.

### 2.1.1 Rubik's Cube

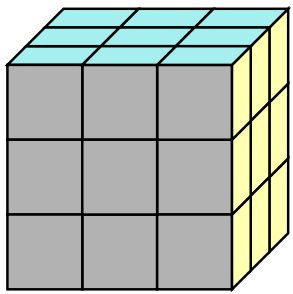

Figure 1: $3 \times 3 \times 3$ Rubik's Cube

Rubik's Cube was invented in 1974 by Erno Rubik of Hungary. The standard version consists of a $3 \times 3 \times 3$ cube (Figure 1), with different colored stickers on each of the exposed squares of the sub-cubes, or cubies. There are 20 movable cubies and 6 stable cubies in the center of each face. The movable cubies can be divided into eight corner cubies, with three faces each, and twelve edge cubies, with two faces each. Corner cubies can only move among corner positions, and edge cubies can only move among edge positions.

Each one of the 6 faces of the cube can be rotated 90, 180, or 270 degrees relative to the rest of the cube. This results in 18 possible moves for each state. Since twisting the same face twice in a row is redundant, the branching factor after the first move can be reduced to 15. In addition, movements of opposite faces are independent. For example, twisting the left face and then the right face leads to the same state as performing the same moves in the opposite order. Pruning redundant moves results in a search tree with an asymptotic branching factor of about 13.34847 (Korf, 1997).

In the goal state, all the squares on each side of the cube are the same color. The puzzle is scrambled by making a number of random moves, and the task is to restore the cube to its original unscrambled state. There are about $4 \times 10^{19}$ different reachable states.

### 2.1.2 The Sliding-tile Puzzles

The sliding-tile puzzle consists of a square frame containing a set of numbered square tiles, and an empty position called the blank. The legal operators are to slide any tile that is horizontally or vertically adjacent to the blank into the blank position. The problem is to rearrange the tiles from some random initial configuration into a particular desired goal configuration. The state space grows exponentially in size as the number of tiles increases, and it has been shown that finding optimal solutions to the sliding-tile problem is NP-complete (Ratner & Warmuth, 1986). The two most common versions of the sliding-tile puzzle are the $3 \times 3$ 8-puzzle, and the $4 \times 4$ 15-puzzle. The 8-puzzle contains 9!/2 (181,440) reachable states, and the 15-puzzle contains about $10^{13}$ reachable states. The goal states of these puzzles are shown in Figure 2.

Figure 2: The 8-puzzle and 15-puzzle goal states

The classic heuristic function for the sliding-tile puzzles is called the Manhattan Distance. It is computed by counting the number of grid units that each tile is displaced from its goal position, and summing these values over all tiles, excluding the blank. Since each tile must move at least its Manhattan Distance to its goal position, and each move changes the location of one tile by just one grid unit, the Manhattan Distance is a lower bound of the minimum number of moves needed to solve a problem instance.

## 2.2 Iterative Deepening A*

Iterative deepening A* (IDA*) (Korf, 1985) performs a series of depth-first searches, increasing a cost threshold $d$ each time. In the depth-first search, all nodes $n$ with $f(n) \leq d$ are expanded. Threshold $d$ is initially set to $h(s)$, where $s$ is the start node. If a goal is found using the current threshold, the search ends successfully. Otherwise, IDA* proceeds to the next iteration by increasing $d$ to the minimum $f$ value that exceeded $d$ in the previous iteration.

## 2.3 Pattern Databases (PDBs)

A powerful approach for obtaining admissible heuristics is to create a simplified version, or abstraction, of the given state space and then to use exact distances in the abstract space as estimates of the distances in the original state space. The type of abstractions we use in this paper for the sliding-tile puzzles are illustrated in Figure 3. The left side of the figure shows a 15-puzzle state $S$ and the goal state. The right side shows the corresponding abstract states, which are defined by erasing the numbers on all the tiles except for 2, 3, 6 and 7. To estimate the distance from $S$ to the goal state in the 15-puzzle, we calculate the exact distance from the abstract state corresponding to $S$ to the abstract goal state.

A *pattern database* (PDB) is a lookup table that stores the distance to the abstract goal of every abstract state (or "pattern") (Culberson & Schaeffer, 1994, 1998). A PDB is built by running a breadth-first search[2] backwards from the abstract goal until the whole abstract space is spanned. To compute $h(s)$ for a state $s$ in the original space, $s$ is mapped to the corresponding abstract state $p$ and the distance-to-goal for $p$ is looked up in the PDB.

---

2. This description assumes all operators have the same cost. This technique can be easily extended to cases where operators have different costs.

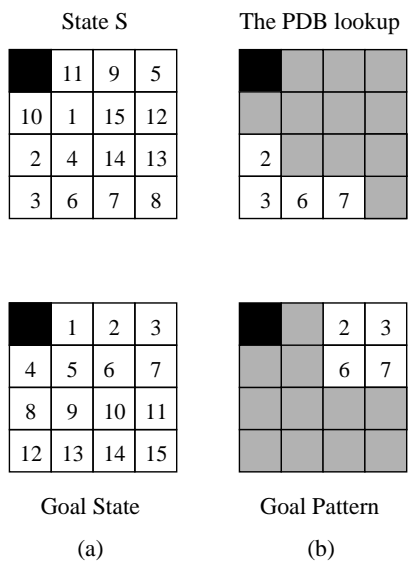

Figure 3: Example of regular lookups

For example, a PDB for the 15-puzzle based on tiles 2, 3, 6, and 7 would contain an entry for every possible way of placing those four tiles and the blank in the 16 puzzle positions. Such a PDB could be implemented as a 5-dimensional array, $PDB$, with the array indexes being the locations of the blank and tiles 2, 3, 6, and 7 respectively. The lookup for state S shown in Figure 3 would then be $PDB[0][8][12][13][14]$ (the blank is in position 0, tile 2 is in position 8, tile 3 is in position 12, etc.). In this paper, accessing the PDB for a state $S$ as just described will be referred to as a *regular* lookup, and the heuristic value returned by a regular lookup will be referred to as a *regular* heuristic value.

*Pattern databases* have proven very useful for finding lower bounds for combinatorial puzzles (Korf, 1997; Culberson & Schaeffer, 1998; Korf & Felner, 2002; Felner, Korf, & Hanan, 2004; Felner et al., 2007). Furthermore, they have also proven to be useful for other search problems, e.g., multiple sequence alignment (McNaughton, Lu, Schaeffer, & Szafron, 2002; Zhou & Hansen, 2004) and planning (Edelkamp, 2001a).

## 2.4 Geometric Symmetries

It is common practice to exploit special properties of a state space to enable additional heuristic evaluations. In particular, additional PDB lookups can be performed given a single PDB. For example, consider Rubik's Cube and suppose we had a PDB based on the positions of the cubies that have a yellow face (the positions of the other cubies don't matter). Reflecting and rotating the puzzle will enable similar lookups for cubies with a different color (*e.g.*, green, red, etc.) since the puzzle is perfectly symmetric with respect to color. Thus, there are 24 symmetric lookups for such a PDB and different heuristic values are obtained for each of these lookups in the same PDB. All these heuristic values are admissible for any given state of the puzzle.

As another example, consider the sliding-tile puzzle. A line of symmetry is the main diagonal (assuming the goal location of the blank is in the upper left corner). Any configuration of tiles can be reflected about the main diagonal and the reflected configuration shares the same attributes as the original one. Such reflections are usually used when using PDBs for the sliding-tile puzzle (Culberson & Schaeffer, 1998; Korf & Felner, 2002; Felner et al., 2004, 2007) and can be looked up from the same PDB.

## 2.5 Methods for Creating Inconsistent Heuristics

With consistent heuristics, the difference between the heuristic value of neighboring nodes is constrained to be *less than or equal to* the cost of the connecting edge. For inconsistent heuristics, there is no constraint on the difference between heuristic values of neighboring nodes and it can be much larger than the cost of the edge connecting them.

The `KRE` formula is designed to work with consistent heuristics and therefore the `KRE` papers report on experiments done with consistent heuristics only. By contrast, our new formula, `CDP`, works for all types of heuristics including inconsistent heuristics. Therefore, in this paper, in addition to the usual consistent heuristics such as regular PDB lookups or Manhattan Distance we also experiment with inconsistent heuristics. We have previously described several methods for producing inconsistent heuristics (Zahavi et al., 2007). Two inconsistent heuristics that are used in the experiments below are the *Random selection of heuristics* and *Dual* evaluations.

- **Random selection of heuristics:** A well-known method for overcoming the pitfalls of a given heuristic is to employ several heuristics and use their maximum value (Holte, Felner, Newton, Meshulam, & Furcy, 2006). For example, multiple heuristics can be based on domain-specific geometric symmetries such as the ones described above. When using geometric symmetries there are no additional storage costs associated with these extra evaluations, even when these evaluations are based on PDBs.

  Although using multiple heuristics results in an improved heuristic value, and therefore is likely to reduce the number of nodes expanded in finding a solution, it also increases the time required to calculate the heuristic values of the nodes, which might increase the overall running time of the search. Instead of using all the available heuristics for every heuristic calculation, one could instead choose to consult only one of them, with the selection being made either randomly or systematically. Because only one heuristic is consulted at each node, the time-per-node is virtually the same as if only one heuristic was available. Even if each of the individual heuristics is consistent, the heuristic values that are *actually used* are inconsistent because different heuristics are consulted at different nodes. We showed (Zahavi et al., 2007) that this inconsistency generally reduces the number of expanded nodes compared to using the same heuristic for all the nodes and it is almost as low as if the maximum over all the heuristics had been computed at every node. For Rubik's Cube, we randomly chose one of the 24 different lookups of the same PDB that arise from the 24 lines of symmetry of this cube.

- **Dual evaluation:** In permutation state spaces such as Rubik's Cube, for each state $s$ there exists a dual state $s^d$ located the same distance from the goal as $s$ (Felner et al., 2005; Zahavi, Felner, Holte, & Schaeffer, 2006; Zahavi et al., 2008). Therefore, any admissible heuristic applied to $s^d$ is also admissible for $s$. The puzzles studied in this paper are permutation state spaces, and the dual of a state in these puzzles is calculated by reversing the role of locations and objects: the "regular" state uses a set of objects indexed by their current location, while the "dual" state has a set of locations indexed by the objects they contain. When using PDBs, a dual lookup is to look up $s^d$ in the PDB. Performing only regular PDB lookups for the states generated during the search produces consistent values. However, the values produced by performing the dual lookup can be inconsistent because the identity of the objects being queried can change dramatically between two consecutive lookups. Due to its diversity, the dual heuristic was shown to be preferable to a regular heuristic (Zahavi et al., 2007). An exact definition and explanations about the dual lookup is provided in the original papers (Felner et al., 2005; Zahavi et al., 2006, 2008).

It is important to note that all three PDB lookups (regular, dual, and random) consult the same PDB. Thus, they need the same amount of memory and share the same overall distribution of heuristic values (Zahavi et al., 2007).

## 3. The KRE Formula and its Limitations

This section begins with a short derivation of the KRE formula for state spaces in which all state transitions have a cost of 1. KRE describe how this can be generalized to account for variable edge costs (Korf et al., 2001).

### 3.1 The KRE formula

For a given state $s$ and IDA* threshold $d$, KRE aims to predict $N(s, d)$, the number of nodes that IDA* will expand if it uses $s$ as its start state and does a complete search with an IDA* threshold of $d$ (*i.e.,* searches to depth $d$ and does not terminate its search if the goal is encountered). This can be written as

$$N(s, d) = \sum_{i=0}^{d} N_i(s, d) \tag{1}$$

where $N_i(s, d)$ is the number of nodes expanded by IDA* at level $i$ when its threshold is $d$. One way to decompose $N_i(s, d)$ is as the product of two terms

$$N_i(s, d) = N_i(s) \cdot P_{ex}(s, d, i) \tag{2}$$

where $N_i(s)$ is the number of nodes in level $i$ of $BFS_s^d$, the brute-force search tree (*i.e.,* the tree created by breadth first search without any heuristic pruning) of depth $d$ rooted at start state $s$, and $P_{ex}(s, d, i)$ is the percentage of the nodes in level $i$ of $BFS_s^d$ that are *expanded* by IDA* when its threshold is $d$.

In KRE, $N_i(s)$ is written as $N_i$, *i.e.,* without the dependence on the start state $s$. This is perfectly correct for state spaces with a uniform branching factor $b$, because $N_i(s)$ in such cases is simply $b^i$. For state spaces with a non-uniform but "regular" branching structure,

`KRE` showed how $N_i$ could be computed exactly using recurrence equations that are independent of $s$. However, the base cases of the recurrences in `KRE` do depend on $s$ so their using $N_i$ instead of $N_i(s)$ is reasonable but not strictly correct.

### 3.1.1 Conditions for Node Expansion in IDA*

To understand how $P_{ex}(s, d, i)$ is treated in `KRE`, it is necessary to reflect on the conditions required for node expansion. A node $n$ in level $i$ of $BFS_s^d$ will be expanded by IDA* if it satisfies two conditions:

1. $f(n) = g(n) + h(n)$ must be *less than or equal to $d$*. When all edges have a unit cost, $g(n) = i$ and this condition is equivalent to $h(n) \leq d - i$. We call nodes that satisfy this condition *potential nodes* because they have the potential to be expanded.

2. $n$ must be generated by IDA*, *i.e.*, its parent (at level $i - 1$) must be expanded by IDA*.

`KRE` restricted its analysis to heuristics that are consistent and proved that in this case the second condition is implied by the first condition. In other words, when the given heuristic is consistent, the nodes expanded by IDA* at level $i$ of $BFS_s^d$ for threshold $d$ are exactly the set of potential nodes at level $i$.[3] This observation allows Equation 2 to be rewritten as

$$N_i(s, d) = N_i(s) \cdot P_{POTENTIAL}(s, i, d - i) \tag{3}$$

where $P_{POTENTIAL}(s, i, v)$ is defined as the percentage of nodes at level $i$ of $BFS_s^d$ whose heuristic value is *less than or equal to $v$*.

Note that although $P_{POTENTIAL}(s, i, d - i) = P_{ex}(s, d, i)$ when the given heuristic is consistent, $P_{POTENTIAL}(s, i, d - i)$ overestimates $P_{ex}(s, d, i)$ when the heuristic is inconsistent, sometimes by a very large amount (see Section 3.2.1).

### 3.1.2 Approximating $P_{POTENTIAL}(s, i, v)$

`KRE` use three different approximations to $P_{POTENTIAL}(s, i, v)$. In `KRE`'s theoretical analysis $P_{POTENTIAL}(s, i, v)$ is approximated by the "equilibrium" distribution, which we denote as $P_{EQ}(v)$. It is defined to be "the probability that a node chosen randomly and uniformly among all nodes at a given depth of the brute-force search tree has heuristic value less than or equal to $v$, in the limit of large depth" (Korf et al. 2001, p. 208). `KRE` proved that, in the limit of large $d$,

$$\sum_{i=0}^{d} N_i(s) \cdot P_{EQ}(d - i)$$

would converge to $N(s, d)$ if the given heuristic is consistent. Their final formula (the `KRE` formula) is therefore:

---

3. See section 3.2.1 below for more discussion of the `KRE` formula with consistent heuristics.

$$N(s, d) = \sum_{i=0}^{d} N_i(s) \cdot P_{EQ}(d - i) \qquad (4)$$

`KRE` contrasted the equilibrium distribution with the "overall" distribution, which is defined as "the probability that a state chosen randomly and uniformly from all states in the problem has heuristic value *less than or equal to v*" (p. 207). Unlike the equilibrium distribution, which is defined over a search tree, the overall distribution is a property of the state space. The overall distribution can be directly computed from a pattern database, if just one pattern database is used and each of its entries corresponds to the same number of states in the original state space, or can be approximated, in more complex settings, by computing the heuristic values of a large random sample of states. `KRE` argued that in Rubik's Cube the overall distribution for a heuristic defined by a single pattern database is the same as the equilibrium distribution, but that for the sliding-tile puzzles, the two distributions are different.

The heuristic used in `KRE`'s experiments with Rubik's Cube was defined as the maximum over three pattern databases. For each individual pattern database, the overall distribution was computed exactly. In `KRE`'s experiments these distributions were combined to approximate $P_{POTENTIAL}(s, i, v)$ by assuming that the values from the three pattern databases were independent.

For the experiments with the sliding-tile puzzles, `KRE` defined three types of states based on the whether the blank was located in a corner position, an edge position, or an interior position, and approximated $P_{POTENTIAL}(s, i, v)$ by a weighted combination of the overall distributions of the states of each type. The weights used at level $i$ were the exact percentages of states of the different types at that level.

In our experiments we followed `KRE` precisely and use the overall distribution for individual Rubik's Cube pattern databases and the weighted overall distribution just described for the sliding-tile puzzles. For simplicity, in the reminder of this paper we use the phrase *unconditional heuristic distribution*[4] and the notation $P(v)$ to refer to the probability that a node has a heuristic *less than or equal to v*. We let the exact context determine which distribution $P(v)$ actually denotes, whether it is the equilibrium distribution, the overall distribution, or any other approximation of $P_{POTENTIAL}$ and $P_{ex}$. Likewise we will use $p(v)$ (lower case $p$) to denote $P(v) - P(v - 1)$ (with $p(0) = P(0)$). $p(v)$ is the probability that a state will have a heuristic value of *exactly v* according to the distribution $P$.

## 3.2 Limitations of the `KRE` Formula

The `KRE` formula (Equation 4) has two main shortcomings: (1) its predictions are not accurate if the given heuristic is inconsistent, and (2) even with consistent heuristics its predictions can be inaccurate for individual start states or sets of start states whose heuristic values are not distributed according to the unconditional heuristic distribution, $P(v)$. We now turn to examine each of these in detail.

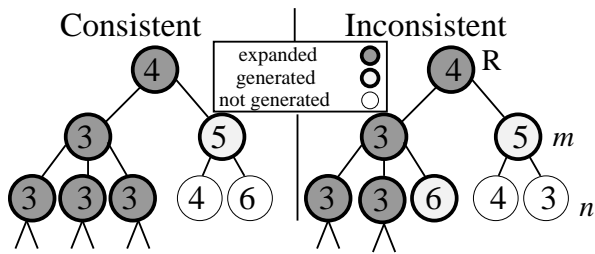

Figure 4: Consistent versus inconsistent heuristics

### 3.2.1 INCONSISTENT HEURISTICS

As specifically mentioned in the KRE papers one property required for the KRE analysis is that the heuristic be *consistent*. This is necessary because the KRE formula aims to count the number of *potential nodes* at each level in $BFS_s^d$. With consistent heuristics, the heuristic value of neighboring states never changes by more than the change in the $g$-value, as illustrated in the left side of Figure 4 (where the number inside a node is its heuristic value). This implies that the $f$-value of a node's ancestor is always less than or equal to the $f$-value of the node (*i.e.*, $f$ is *monotone non-decreasing* along a path in the search tree). Therefore, it is easy to prove that with consistent heuristics all the ancestors of a potential node are also potential nodes (Korf et al., 2001). Consequently IDA* will expand all and only the potential nodes in $BFS_s^d$. Hence, a formula such as KRE that aims to count the number of potential nodes in $BFS_s^d$ can be used to predict the number of nodes IDA* will expand when given a consistent heuristic.

For inconsistent heuristics this reasoning does not apply. The heuristic values of neighboring states can differ by much more than the cost of the edge that connects them, and thus the $f$-values along a path in the search tree are not guaranteed to be monotonically non-decreasing. Therefore, the ancestors of a potential node are not guaranteed to be potential nodes themselves, with the consequence that a potential node might never be generated. For example, consider the search tree in the right side of Figure 4. The numbers inside each node show the node's heuristic value. Assume that the start node is $R$ and that the IDA* threshold is 5 (a node is a potential node if its $f$-value is *less than or equal to* 5). There are 3 potential nodes at depth 2 (all with heuristic value 3). Consider the potential node $n$. The path to it is through node $m$ but node $m$ is not a potential node ($f(m) = 1 + 5 = 6 > 5$), so it will be generated but not expanded. Therefore, node $n$ will never be generated, preventing IDA* from expanding it. Since the KRE formula counts the number of potential nodes, it will count node $n$ and thus overestimate the number of expanded nodes when an inconsistent heuristic is used.

The amount by which KRE overestimates the number of nodes expanded by IDA* with an inconsistent heuristic can be very large. To illustrate this, consider the state space for Rubik's Cube and a PDB heuristic defined by the locations of 6 (out of 12) of the edge cubies. The regular method for looking up a heuristic value in a PDB produces a consistent heuristic. As discussed in Section 2.5 two alternative PDB lookups that produce inconsistent heuristics are the *dual evaluation* and the *random selection* from multiple heuristics. In Rubik's Cube there are 24 symmetries and each can be applied to any state to create a new way to perform a PDB lookup for it. Thus, there are 24 heuristics for Rubik's Cube based on the same PDB and the random-symmetry lookup chooses one of them randomly.

---

4. "unconditional" to distinguish it from the conditional distribution we introduce in Section 4.1

| d | KRE | Regular | Dual | Random Symmetry |
|---|---|---|---|---|
| 8 | 257 | 277 | 36 | 26 |
| 9 | 3,431 | 3,624 | 518 | 346 |
| 10 | 45,801 | 47,546 | 6,809 | 4,608 |
| 11 | 611,385 | 626,792 | 92,094 | 61,617 |
| 12 | 8,161,064 | 8,298,262 | 1,225,538 | 823,003 |
| 13 | 108,937,712 | 110,087,215 | 16,333,931 | 10,907,276 |

Table 1: Rubik's Cube - Number of nodes expanded by IDA* using regular, dual, and random-symmetry PDB lookups for different IDA* threshold $d$ and the corresponding KRE predictions.

Because all three lookups (regular, dual, and random-symmetry) consult the same PDB they have the same distribution of heuristic values, $P(v)$, and therefore KRE will predict that IDA* will expand the same number of nodes regardless of whether a regular, dual, or random-symmetry lookup is done. The experimental results in Table 1 show that a substantially different number of nodes are actually expanded in practice for each of these methods.

Each row of Table 1 presents results for a specific IDA* threshold ($d$). Each result is an average over $1,000$ random initial states, each of which is generated by making 180 random moves from the goal state. The "KRE" column shows the KRE prediction based on the unconditional heuristic distribution. The last three columns of Table 1 show the number of nodes IDA* expands when it performs either a regular, dual, or random-symmetry lookup in the PDB. The KRE prediction is within 8% of the actual number of nodes expanded when IDA* uses the regular (consistent) PDB lookup (third column) but it substantially overestimates the number of nodes expanded when IDA* uses the dual or random-symmetry inconsistent lookups in the same PDB (fourth and fifth columns).

### 3.2.2 Sets of Start States Whose Heuristics Values do not Obey the unconditional heuristic distribution

As explained above, KRE used the unconditional heuristic distribution $P(v)$ and, in their theoretical analysis, proved that its use in the KRE formula would give accurate predictions in the limit of large depth. In fact, accurate predictions will occur as soon as the heuristic distribution at the depth of interest $d$ closely approximates $P(v)$. This happens at large depths by definition but this can happen even at very shallow levels under certain circumstances. The reason that KRE was able to produce extremely accurate predictions in its experiments using the unconditional heuristic distribution $P(v)$ for all depths and all start states is that its experiments report average predictions and performances over a large number of randomly drawn start states. In the spaces used in `KRE`'s experiments, the heuristic distribution of a large random set of start states very closely approximated the $P(v)$ distribution they used. This caused heuristic distributions at all levels to closely approximate $P(v)$.

However, if the set of start states does not have its heuristic values distributed according to $P(v)$, as is the case for most non-random sets of start states or a single start state, `KRE` should not be expected to make good predictions for small depths. In other words, in such cases the unconditional heuristic distribution $P(v)$ is not expected to be a good approximation of $P_{ex}(s, d, i)$.

Consider the case of a single start state and a consistent heuristic. The distribution of heuristic values in the search tree close to the start state will be highly correlated with the heuristic value of the start state, and therefore will not be the same in search trees with start states having different heuristic values. For example, a great deal of pruning is likely to occur near the top of the search tree for a start state with a large heuristic value, resulting in fewer nodes expanded than for a start state with a small heuristic value. Applying `KRE` to these two states will produce the same prediction, and therefore be inaccurate for at least one of them, because it uses the same unconditional heuristic distribution $P(v)$ in both cases.

| $h$ | IDA* | `KRE` |
|---|---|---|
| 5 | 30,363,829 | 8,161,064 |
| 6 | 18,533,503 | 8,161,064 |
| 7 | 10,065,838 | 8,161,064 |
| 8 | 6,002,025 | 8,161,064 |
| 9 | 3,538,964 | 8,161,064 |

Table 2: Results for a set of 1,000 start states all with the $h$-value shown in the first column (regular PDB lookup, IDA* threshold of $d = 12$)

Table 2 demonstrates this phenomenon on Rubik's Cube with one regular 6-edge PDB lookup for IDA* threshold $d = 12$. The "IDA*" column shows the average number of nodes expanded for 1,000 start states, all with the *same* heuristic value $h$ for any given row. `KRE` ignores the heuristic values of the start states and predicts that 8,161,064 nodes will be expanded by IDA* for every start state. The row for $d = 12$ in Table 1 shows that this is an accurate prediction when performance is averaged over a large random sample of start states, but in Table 2 we see that it is too low for start states with small heuristic values and too high for ones with large heuristic values.

### 3.2.3 Convergence of the Heuristic Distributions at Large Depths

As described above, `KRE` will make accurate predictions for level $i$ if the nodes at that level actually obey the unconditional heuristic distribution $P(v)$. As $i$ increases, the distribution of heuristic values will start to *converge to* $P(v)$. The rate of convergence depends upon the state space. It is believed to be fairly slow for the sliding-tile puzzles, but faster for Rubik's Cube. If the convergence occurs before the IDA* threshold is reached `KRE` will provide accurate predictions for any set of start states (including single start states).

In order to experimentally test this we repeated the `KRE` Rubik's Cube experiment but, in addition to using a large set of random start states, we also looked at the individual performance of two start states, $s_6$, which has a low heuristic value (6), and $s_{11}$, which has the maximum value for the heuristic used in this experiment (11). As in `KRE` we used the 8-6-6 heuristic which takes the maximum of 3 different PDBs (one based on all 8 corner cubies and two based on 6 edge cubies each). This heuristic is admissible and consistent. Over the billion random states we sampled to estimate $P(v)$ the maximum value was 11 and the average value was 8.898.

| | KRE | Multiple start states | | Single start state | | | |
|---|---|---|---|---|---|---|---|
| $d$ | | IDA* | Ratio | $s_6$ | Ratio | $s_{11}$ | Ratio |
| 10 | 1,510 | 1,501 | 0.99 | 53,262 | 0.03 | - | - |
| 11 | 20,169 | 20,151 | 1.00 | 422,256 | 0.05 | 8,526 | 2.37 |
| 12 | 269,229 | 270,396 | 1.00 | 3,413,547 | 0.08 | 162,627 | 1.66 |
| 13 | 3,593,800 | 3,564,495 | 0.99 | 29,114,115 | 0.12 | 2,602,029 | 1.38 |
| 14 | 47,971,732 | 47,961,699 | 1.00 | 259,577,913 | 0.18 | 38,169,381 | 1.26 |
| 15 | 640,349,193 | 642,403,155 | 1.00 | 2,451,954,240 | 0.26 | 542,241,315 | 1.18 |
| 16 | 8,547,681,506 | 8,599,849,255 | 1.01 | 24,484,797,237 | 0.35 | 7,551,612,957 | 1.13 |
| 17 | 114,098,463,567 | 114,773,120,996 | 1.01 | 258,031,139,364 | 0.44 | 103,934,322,960 | 1.10 |

Table 3: Rubik's Cube - Max of (8,6,6) PDBs

Table 3 presents the results. The `KRE` column presents the `KRE` prediction and the *Multiple start states* columns presents the actual number of states generated (averaged over a set of random start states) for each IDA* threshold. Both columns are copied from the `KRE` journal paper (Korf et al., 2001). The Ratio columns of Table 3 shows the value predicted by the `KRE` formula divided by the actual number of nodes generated. This ratio was found to be very close 1.0 for multiple start states, indicating that `KRE`'s predictions were very accurate.

The results for the two individual start states we tested are shown in the "Single start state" part of the table. Note that both states are optimally solved at depth 17, but, as in `KRE`, the search at that depth was run to completion. In both cases the `KRE` formula was not accurate for small thresholds but the accuracy of the prediction increased as the threshold increased. At threshold $d = 17$ the `KRE` prediction was roughly a factor of 2 too small for $s_6$ and about 10% too large for $s_{11}$. This is a large improvement over the smaller thresholds. These predictions will become even more accurate as depth continues to increase.

The reason the predictions improve for larger values of $d$ is that at deeper depths the heuristic distribution within a single level converges to the unconditional heuristic distribution. Using dashed and dotted lines of various types, Figure 5(a) shows the distribution of heuristic values seen in states that are 0, 1, 2 and 4 moves away from $s_6$. The solid line in Figure 5(a) is the unconditional heuristic distribution. The $x$-axis corresponds to different heuristic values and the $y$-axis shows the percentage of states at the specified depth with heuristic values less than or equal to each $x$ value. For example for depth 0 (which includes the start state only) only a heuristic value of 6 was seen (leftmost curve). For depth 1, heuristic values of 5, 6 and 7 were seen (second curve from the left), and so on. The figure shows that the heuristic distribution at successive depths converges to the unconditional heuristic distribution (rightmost curve in Figure 5(a)). At depth 17 (not shown), the heuristic distribution is probably quite close to the unconditional heuristic distribution, making the KRE prediction quite accurate even for this single start state.

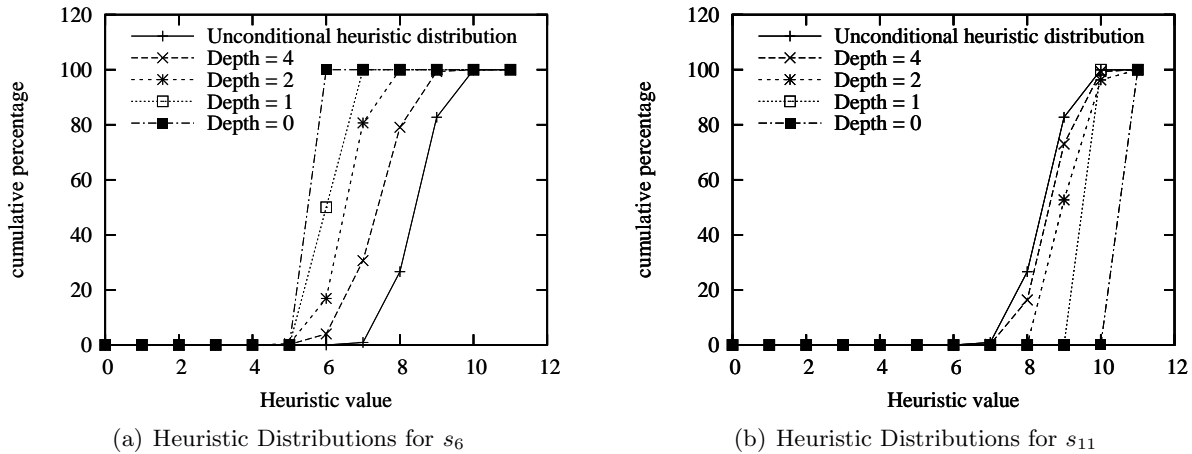

(a) Heuristic Distributions for $s_6$    (b) Heuristic Distributions for $s_{11}$

Figure 5: Convergence of heuristic distributions

Figure 5(b) shows the heuristic distributions for nodes that are 0, 1, 2, and 4 moves away from $s_{11}$. In this case the unconditional heuristic distribution is to the left of the heuristic distributions for the shallow depths, with the heuristic distribution for depth 0 being the rightmost curve in this figure. Comparing parts (a) and (b) of Figure 5 we see that the convergence to the unconditional heuristic distribution is faster for $s_{11}$ than for $s_6$, which explains why the KRE prediction in Table 3 is more accurate for $s_{11}$.

## 4. Conditional Distribution and the CDP Formula

We now present our new formula CDP (Conditional Distribution Prediction), which overcomes the two shortcomings of KRE described in the previous section. An important feature of CDP is that it extends the unconditional heuristic distribution of heuristic values $P(v)$ used in KRE to be a *conditional distribution*.

### 4.1 Conditional Distribution of Heuristic Values

The *conditional distribution* of heuristic values is denoted $P(v|context)$, where the *context* represents local properties of the search tree in the neighborhood of a node that influence the distribution of heuristic values in the node's children. Specifically, if $P_n(v)$ is the percentage of node $n$'s children that have a heuristic value *less than or equal to* $v$, then we define $P(v|context)$ to be the average of $P_n(v)$ over all nodes $n$ that satisfy the conditions defined by the context. $P(v|context)$ can be interpreted as the probability that a node with heuristic value less than or equal to $v$ will be produced when a node satisfying the conditions specified by *context* is expanded. When the context is empty it is denoted $P(v)$ as in Section 3. We use $p(v|context)$ (lower case $p$) to denote the probability that a node with heuristic value *equal* to $v$ will be produced when a node satisfying the conditions specified by *context* is expanded. Obviously, $P(v|context) = \sum_{i=0}^{v} p(i|context)$.

### 4.1.1 The Basic 1-Step Model

The conditioning *context* can be any combination of local properties of the search tree, including properties of the node itself (*e.g.* its heuristic value), the operator that was applied to generate the node, properties of the node's ancestors in the search tree, etc. The simplest conditional distribution is $p(v|v_p)$, the probability of a node with a heuristic value equal to $v$ being produced when a node with value $v_p$ is expanded. We call this a *1-step* model because each value is conditioned by nodes that are one step away only. In special circumstances, $p(v|v_p)$ can be determined exactly by analysis of the state space and the heuristic, but in general it must be approximated empirically by sampling the state space.

In our sampling method $p(v|v_p)$ is represented by the entry $M[v][v_p]$ in a two-dimensional matrix $M[0..h_{max}][0..h_{max}]$, where $h_{max}$ is the maximum possible heuristic value. To build the matrix we first set all values in the matrix to 0. We then randomly generate a state and calculate its heuristic value $v_p$. After that, we generate each child of this state one at a time, calculate the child's heuristic value ($v$), and increment $M[v][v_p]$. We repeat this process a large number of times in order to generate a large sample. Finally, we divide the value of each cell of the matrix by the sum of the column the cell belongs to, so that entry $M[v][v_p]$ represents the percentage of children generated with value $v$ when a state with value $v_p$ is expanded.

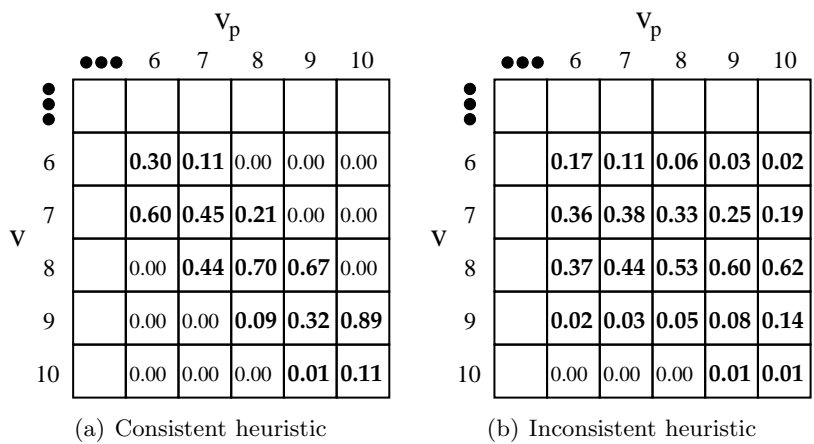

(a) Consistent heuristic         (b) Inconsistent heuristic

Figure 6: A portion of the *Conditional Distribution* matrix for Rubik's Cube for consistent and for inconsistent heuristics

Figure 6 shows the bottom right corner of two such matrices for the 6-edge PDB of Rubik's Cube. The left matrix (a) shows $p(v|v_p)$ for the regular (consistent) lookup in this PDB and the right matrix (b) shows $p(v|v_p)$ for the inconsistent heuristic created by the dual lookup in this PDB. The matrix in (a) is tridiagonal because neighboring values cannot differ by more than 1. For example, states with a heuristic value of 8 can only have children with heuristics of 7, 8 and 9; these occur with probabilities of 0.21, 0.70 and 0.09 respectively (see column 8). By contrast, the matrix in (b) is not tridiagonal. In column 8, for example, we see that 6% of the time states with heuristic value of 8 have children with heuristic values of 6.

### 4.1.2 Richer Models

When IDA* expands a node, it eliminates some children because of operator pruning. For example, in state spaces with undirected operators, such as we are using in our studies, the parent of a node would be generated among the node's children but IDA* would immediately prune it away. Distribution $p(v|v_p)$ does not take this into account. In order to take this into consideration it is necessary to extend the context of the conditional probability to include the heuristic value of the parent of the node being expanded (we refer to the parent node as $gp$). We denote this by $p(v|v_p,v_{gp})$ and call this a "2-step" model because it conditions on information from ancestors up to two steps away. $p(v|v_p,v_{gp})$ gives the probability of a node with a heuristic value equal to $v$ being generated when the node being expanded has a heuristic value of $v_p$ and the parent of the node being expanded has a heuristic value of $v_{gp}$. It is estimated by sampling in the same way as was done to estimate $p(v|v_p)$, except that each sample generates a random state, $gp$, then all its neighbors, and then all of their neighbors except those eliminated by operator pruning. Naturally, the results of the sampling for this 2-step model are stored in a three-dimensional array.

The context of the conditional distribution can be extended in other ways as well. For the sliding-tile puzzles, KRE conditions the overall distribution on the "type" of the state being expanded, where the type indicates if the blank is in a corner, edge, or interior location. In our experiments with the sliding-tile puzzle below, we extend $p(v|v_p,v_{gp})$ with this type information: $p(v,t|v_p,t_p,v_{gp},t_{gp})$ gives the probability of a node of type $t$ with heuristic value equal to $v$ being generated when the node being expanded has heuristic value $v_p$ and type $t_p$ and the expanded node's parent has heuristic value $v_{gp}$ and type $t_{gp}$.

## 4.2 A New Prediction Formula, CDP (Conditional Distribution Prediction)

In this section we use the conditional distributions just described to develop CDP, an alternative to the KRE formula for predicting the number of nodes IDA* will expand for a given heuristic, IDA* threshold, and set of start states. As will be shown below experimentally, the new formula CDP overcomes the limitations of KRE and works well for inconsistent heuristics and for any set of start states with arbitrary IDA* threshold.

Our overall approach is as follows. Define $N_i(s,d,v)$ to be the number of nodes that IDA* will *generate* at level $i$ with a heuristic value equal to $v$ when $s$ is the start state and $d$ is the IDA* threshold. Given $N_i(s,d,v)$, the number of nodes IDA* will expand at level $i$ for threshold $d$ is $\sum_{v=0}^{d-i} N_i(s,d,v)$, and, $N(s,d)$, the total number of nodes expanded in a complete iteration of IDA* with threshold $d$ over all levels, the quantity we are ultimately interested in, is $\sum_{i=0}^{d} \sum_{v=0}^{d-i} N_i(s,d,v)$. In these summations $v$ only runs up to $d-i$ because only nodes with heuristic values in the range $[0 \dots d-i]$ will be expanded at level $i$.

If $N_i(s,d,v)$ could be calculated exactly, this formula would calculate $N(s,d)$ exactly whether the given heuristic is consistent or not. However, there is no general method for efficiently calculating $N_i(s,d,v)$ exactly. Instead, $N_i(s,d,v)$ will be estimated recursively from $N_{i-1}(s,d,v)$ and the conditional distribution; the exact details depend on the conditional model being used and are given in the subsections that follow. We will use $\tilde{N}_i(s,d,v)$ to denote an approximation of $N_i(s,d,v)$. In Section 4.5.1 we will describe conditions in which this calculation is, in fact, exact, and therefore produces perfect predictions of $N(s,d)$. But in the general case these predictions may not be perfect and are only estimates. At the present time we have no analytical tools for estimating their accuracy but as we show experimentally, these estimates are often very accurate.

### 4.3 Prediction Using the Basic 1-Step Model

If the basic 1-step conditional distribution $p(v|v_p)$ is being used, $N_i(s,d,v)$ can be estimated recursively as follows:

$$N_i(s,d,v) \approx \tilde{N}_i(s,d,v) = \sum_{v_p=0}^{d-(i-1)} \tilde{N}_{i-1}(s,d,v_p) \cdot b_{v_p} \cdot p(v|v_p) \tag{5}$$

where $b_{v_p}$ is the average branching factor of nodes with heuristic value $v_p$, which is estimated during the sampling process that estimates the conditional distribution.[5] The reasoning behind this equation is that $N_{i-1}(s,d,v_p) \cdot b_{v_p}$ is the total number of children IDA* generates via the nodes it expands at level $i-1$ with heuristic value equal to $v_p$. This is multiplied by $p(v|v_p)$ to get the expected number of these children that have heuristic value $v$. Nodes at level $i-1$ are expanded if and only if their heuristic value is less than or equal to $d-(i-1)$, hence the summation only includes $v_p$ values in the range of $[0 \dots d-(i-1)]$. By restricting $v_p$ to be less than or equal to $d-(i-1)$ in every recursive application of this formula, we ensure (even for inconsistent heuristics) that a node is only counted at level $i$ if all its ancestors are expanded by IDA*. The base case of this recursion, $N_0(s,d,v)$, is 1 for $v = h(s)$ and 0 for all other values of $v$.

Based on this, the number of nodes expanded by IDA* given start state $s$, threshold $d$, and a particular heuristic can be predicted as follows:

$$\texttt{CDP}_1(s,d) = \sum_{i=0}^{d} \sum_{v=0}^{d-i} \tilde{N}_i(s,d,v) \tag{6}$$

If a set, $S$, of start states is given instead of just one start state, the calculation is identical except that the base case of the recursion is defined using all the start states in $S$. That is, we define $N_0(S,d,v)$ to be equal to $k$ if there are $k$ states in $S$ with a heuristic value of $v$. The rest of the formula remains the same (with $S$ substituted for $s$ everywhere).

---

5. In the general case of our equation the branching factor depends on the *context* that defines the conditional distribution. Since in the 1-step model, the context is just the heuristic value $v$, we formally allow the branching factor to depend on it. In practice, the branching factor is usually the same for all heuristic values.

### 4.4 Prediction Using Richer Models

If the 2-step conditional distribution $p(v|v_p, v_{gp})$ is being used, we define $N_i(s, d, v, v_p)$ to be the number of nodes that IDA* will *generate* at level $i$ with a heuristic value equal to $v$ from nodes at level $i - 1$ with heuristic value $v_p$ when $s$ is the start state and $d$ is the IDA* threshold. $N_i(s, d, v, v_p)$ can be estimated recursively as follows:

$$N_i(s, d, v, v_p) \approx \tilde{N}_i(s, d, v, v_p) = \sum_{v_{gp}=0}^{d-(i-2)} \tilde{N}_{i-1}(s, d, v_p, v_{gp}) \cdot b_{v_p, v_{gp}} \cdot p(v|v_p, v_{gp}) \qquad (7)$$

where $b_{v_p, v_{gp}}$ is the average branching factor of nodes with heuristic value $v_p$ and a parent with heuristic value $v_{gp}$. The base case for this 2-step model is at level 1, not level 0. $N_1(s, d, v, v_p)$ is 0 for $v_p \neq h(s)$, and is the number of children of the start state $s$ with heuristic value $v$ for $v_p = h(s)$. Based on this 2-step model the number of nodes expanded by IDA* given start state $s$, threshold $d$, and a particular heuristic can be predicted as follows:

$$\mathtt{CDP}_2(s, d) = \sum_{i=0}^{d} \sum_{v=0}^{d-i} \sum_{v_p=0}^{d-(i-1)} \tilde{N}_i(s, d, v, v_p) \qquad (8)$$

If there is a set $S$ of start states instead of just one, the base case is $N_1(S, d, v, v_p)$, the number of children with heuristic value $v$ of the states in $S$ with heuristic value $v_p$.

Analogous definitions of $N_i$ and $\mathtt{CDP}$ can be used with any definition of context. For example, if using a 1-step model with a set $T$ of state types, one would define $N_i(s, d, v, t)$ as the number of nodes of type $t$ that IDA* will *generate* at level $i$ with a heuristic value equal to $v$, and estimate it recursively as follows:

$$N_i(s, d, v, t) \approx \tilde{N}_i(s, d, v, t) = \sum_{v_p=0}^{d-(i-1)} \sum_{t_p \in T} \tilde{N}_{i-1}(s, d, v_p, t_p) \cdot b_{v_p, t_p} \cdot p(v, t|v_p, t_p) \qquad (9)$$

Based on this model the number of nodes expanded by IDA* given start state $s$, threshold $d$, and a particular heuristic can be predicted as follows:

$$\mathtt{CDP}(s, d) = \sum_{i=0}^{d} \sum_{v=0}^{d-i} \sum_{t \in T} \tilde{N}_i(s, d, v, t) \qquad (10)$$

### 4.5 Prediction Accuracy

The accuracy of our predictions can be arbitrarily good or arbitrarily bad depending on the accuracy of the conditional model being used. In the following subsections we examine each of these extreme cases.

In principle, extending the context should never decrease the accuracy of the predictions because additional information is taken into account. However, when the conditional model is being estimated by sampling, an extended context can result in poorer predictions because there are fewer samples in each context. This is our explanation of why the 1-step model is more accurate than the 2-step model in rows $h = 6$ and $h = 9$ in Table 7 in Section 6.2 below.

### 4.5.1 Perfect Predictions

Consider any definition of context that includes the heuristic value of the node being expanded ($v_p$ in the contexts defined above) and contains sufficient information to allow operator pruning to be correctly accounted for. We will use the notation $(v, x)$ to refer to a specific instance of such a context, where $v$ is the heuristic value of the node being expanded and $x$ is an instantiation of the other information in the the context (*e.g.*, the state type information in the last model above). The general form of our predictive model with such a context is

$$\text{CDP}(s, d) = \sum_{i=0}^{d} \sum_{v=0}^{d-i} \sum_{\substack{\text{all } x \\ \text{such that } (v,x) \\ \text{is an instance} \\ \text{of the context}}} \tilde{N}_i(s, d, v, x) \tag{11}$$

with

$$\tilde{N}_i(s, d, v, x) = \sum_{v_p=0}^{d-(i-1)} \sum_{\substack{\text{all } x_p \\ \text{such that } (v_p, x_p) \\ \text{is an instance} \\ \text{of the context}}} \tilde{N}_{i-1}(s, d, v_p, x_p) \cdot b_{v_p, x_p} \cdot p(v, x | v_p, x_p) \tag{12}$$

where $b_{v_p, x_p}$ is the average branching factor, after operator pruning, of all nodes satisfying the conditions of context $(v_p, x_p)$, and $p(v, x | v_p, x_p)$ is the average over all nodes $n$ satisfying the conditions of context $(v_p, x_p)$ of $p_n(v, x)$, the percentage of $n$'s children, after operator pruning, that satisfy the conditions of context $(v, x)$.

If, for every context $(v_p, x_p)$, all nodes $n$ satisfying the conditions defined by $(v_p, x_p)$ have exactly the same branching factor $b_{v_p, x_p}$ and exactly the same value of $p_n(v, x)$ for all contexts $(v, x)$, a simple proof by induction starting from the correctness of the base cases, $N_1(s, d, v, x)$, shows that $N_i(s, d, v, x) = \tilde{N}_i(s, d, v, x)$ for all $i$, i.e., that our prediction method correctly calculates exactly how many nodes will satisfy the conditions of each context at every level of the search tree. From this it follows that $\text{CDP}(s, d)$ will be exactly the number of nodes IDA* will expand given start state $s$ and IDA* threshold $d$.

A practical setting in which the predictions of our 2-step model are guaranteed to be perfect by this reasoning is when the following conditions hold:

1. the heuristic is defined to be the exact distance to the goal in an abstract state space, as is the case when a single pattern database is used.

2. any two states, $s_1, s_2$, that map to the same abstract state $x$ have the same set of operators $\{op_1, ..., op_k\}$ that apply to them, and

3. if states $s_1$ and $s_2$ map to abstract state $x$, then for all operators $op \in \{op_1, ..., op_k\}$ that apply to $s_1$ and $s_2$, $s_1$'s child $op(s_1)$ and $s_2$'s child $op(s_2)$ map to the same abstract state, $op(x)$.

Define the context of a node to be its heuristic value and the abstract state to which it maps. Condition (2) guarantees that for every context $(v, x)$, all nodes satisfying the conditions of $(v, x)$ have exactly the same branching factor $b_{v,x}$. This is true because if nodes $n_1$ and $n_2$ satisfy the conditions of context $(v, x)$, they both map to the same abstract state $x$, and condition (2) then requires that exactly the same set of operators apply to them both. Conditions (2) and (3) together guarantee that for every context $(v_p, x_p)$, all nodes satisfying the conditions of $(v_p, x_p)$ have exactly the same value of $p_n(v, x)$ for all $v$ and $x$. This is true because if nodes $n_1$ and $n_2$ satisfy the conditions of context $(v_p, x_p)$, they both map to the abstract state $x_p$, the same set of operators applies to both, and each operator $op$ creates a child that, in both cases, maps to a specific abstract state, $op(x_p)$. Therefore the percentage of children that map to any particular abstract state is the same for both $n_1$ and $n_2$.

A straightforward implementation of the prediction method in this setting associates a counter with each abstract state, which is initialized to the number of start states that map to the abstract state. The counter for abstract state $x$ is updated once for each value of $i$ ($1 \le i \le d$) by adding to it, for each operator $op$, the current value of the counter of each abstract state $y$ such that $op(y) = x$. This algorithm has a computational complexity that is $O(d \times |A| \times \beta^2)$ where $|A|$ is the number of abstract states and $\beta$ is the effective branching factor in the abstract space. Because the complexity depends only linearly on $d$, in contrast to the typically exponential dependency on $d$ for the number of nodes IDA* will expand, for sufficiently large $d$ the prediction will be arbitrarily faster to compute than the search itself. For example, for a PDB for the 15-puzzle based on the positions of 8 tiles and the blank (roughly 4 billion abstract states), the prediction for 1000 start states with $d = 52$ takes only 6% of the time required to execute the search.

An exact prediction for this setting has two potential uses. The first is to determine if searching with a single PDB is feasible or not. For example, the calculation might show that even the first iteration of IDA* (with a threshold of $h(start)$) will take more than a year to complete. The second is to use the prediction to compare the actual performance of an alternative method executed on a set of start states (*e.g.* taking the maximum over a set of PDBs) to the performance using a single PDB without actually having to execute the IDA* search with the single PDB.

### 4.5.2 VERY POOR PREDICTIONS

The predictions made by a conditional model will be extremely inaccurate if the distribution of heuristic values is independent of the information supplied by the context. We illustrate this with an example based on the 4x3 sliding-tile puzzle and two heuristics, a PDB based on the locations of tiles 1-7 and the blank, and the heuristic that returns 0 for every state. If the given state has the blank in its goal position, or in a position that is an even number of moves from the goal position, the heuristic value for that state is taken from the PDB. The other states have a heuristic value of 0. In the search tree, the heuristic used at level $i$ will therefore be the opposite of the one used at level $i - 1$.

A 1-step model in this situation will clearly be hopeless for predicting the heuristic distribution in levels where the PDB is being used until $i$ is sufficiently large that the distribution at level $i$ converges to the unconditional distribution.

There is some hope that a 2-step model could make reasonably accurate predictions because the PDB, considered by itself, defines a consistent heuristic and therefore the distribution of heuristic values of a node's children are somewhat correlated with the heuristic value of the node's parent.

We tested this using the 4x3 sliding-tile puzzle, which is small enough that we could build our 2-step model using all the states in the state space so that no error is introduced through a sampling process. To test the prediction accuracy of the model we generated 50,000 solvable states at random and, as will be explained in detail in the next section, used state $s$ as a start state in combination with IDA* threshold $d$ if IDA* would actually have executed an iteration with threshold $d$ when given state $s$ as a start state. This means that a different number of start states might be used for each value of $d$. The "Num" column in Table 4 indicates how many start states were used for each value of $d$ (first column) and we have only included in this table results for which more than 5,000 start states were used. The "IDA*" column shows the average number of nodes expanded by IDA* on the start states used for each $d$ and the "Prediction" column shows the number predicted by our 2-step model. The "Ratio" column is "Prediction" divided by "IDA*". One can clearly see the improvement of the predictions as $d$ increases. But even at the deepest depth at which our sample provided more than 5,000 start states, the prediction is a factor of 6 smaller than the true value. Of course, using the constant heuristic value of 0 in alternate levels is not something one would do in practice, but we obtained similar results, for essentially the same reason, with the 15-puzzle when switching, from one level to the next, between a pattern database based on tiles 1-7 and a pattern database based on tiles 9-15 (see Section 7.1).

| d | IDA* | CDP$_2$ | | Num |
|---|---|---|---|---|
| | | Prediction | Ratio | |
| 27 | 1,212 | 48 | 0.04 | 5,754 |
| 28 | 1,529 | 63 | 0.04 | 7,780 |
| 29 | 2,340 | 90 | 0.04 | 9,086 |
| 30 | 3,072 | 131 | 0.04 | 11,561 |
| 31 | 4,818 | 208 | 0.04 | 12,397 |
| 32 | 6,607 | 338 | 0.05 | 14,109 |
| 33 | 10,748 | 585 | 0.05 | 14,109 |
| 34 | 15,184 | 1,027 | 0.07 | 14,545 |
| 35 | 24,613 | 1,896 | 0.08 | 13,492 |
| 36 | 36,726 | 3,513 | 0.10 | 12,261 |
| 37 | 60,779 | 6,737 | 0.11 | 10,405 |
| 38 | 96,077 | 12,941 | 0.13 | 8,355 |
| 39 | 152,079 | 25,119 | 0.17 | 6,505 |

Table 4: 4x3 sliding-tile puzzle, alternating between a good heuristic and 0.

## 5. Experimental Setup

The next two sections describe the experimental results that we obtained by running IDA* and comparing the number of nodes it expanded to the number predicted by `KRE` and by

`CDP`. We experimented on the same two application domains used by `KRE`, namely, Rubik's Cube (Section 6) and the sliding-tile puzzle (Section 7). In each domain we evaluated the accuracy of the two formulas, for both consistent and inconsistent heuristics, on a set of solvable start states that were generated at random.

In all the experiments reported here, the start states used for a given IDA* threshold $d$ were subject to a special condition. State $s$ is only used as a start state in combination with threshold $d$ if IDA* actually performs an iteration with threshold $d$ when $s$ is the start state. For example, we would not use $s$ as a start state for $d = 17$ if $s$ is only distance 11 from the goal or if $h(s) > 17$. In addition, for the sliding-tile puzzle, start state $s$ would not be used with IDA* threshold $d$ if $h(s)$ and $d$ were of different parity. By contrast, the experiments in the `KRE` paper did not restrict the choice of start states in this way, the same start states were used with every IDA* threshold .

This difference in how start states are chosen can have a large impact on the number of nodes IDA* expands. Table 5 illustrates this for the 15-Puzzle using the Manhattan Distance heuristic for IDA* threshold $d$ (first column) between 43 and 50. The "nodes" column under "Unrestricted" shows the number of nodes IDA* expanded on average for 50,000 randomly generated solvable start states. The values in this column are in close agreement with the corresponding results of Table 5 in the `KRE` paper (Korf et al., 2001). The "number" column shows how many of these start states satisfy our additional condition. If we remove the start states that violate our condition, IDA* expands substantially fewer nodes on average, as shown in the "nodes" column under "Restricted", and the difference increases as $d$ increases. At $d = 50$ there is almost an order of magnitude difference between the number of nodes expanded in the two settings. This difference needs to be kept in mind when making comparisons with the experimental results reported here and in the `KRE` papers.

| | Unrestricted | Restricted | |
|---|---|---|---|
| $d$ | nodes | number | nodes |
| 43 | 439,942 | 22,525 | 219,001 |
| 44 | 1,014,941 | 22,484 | 393,406 |
| 45 | 1,985,565 | 22,937 | 688,119 |
| 46 | 4,542,249 | 22,266 | 1,182,522 |
| 47 | 8,963,747 | 22,243 | 2,108,766 |
| 48 | 20,355,110 | 21,028 | 3,508,482 |
| 49 | 40,479,725 | 20,389 | 6,037,064 |
| 50 | 91,329,281 | 18,758 | 9,904,973 |

Table 5: 15-Puzzle with Manhattan Distance. The effect on nodes expanded if start states are randomly chosen or subject to our condition.

## 6. Experimental Results for Rubik's Cube

We begin with Rubik's Cube experiments. The heuristic used here is the 6-edge PDB heuristic described above (Section 3.2.1). We experimented with the (consistent) regular lookup and the (inconsistent) random-symmetry and dual lookups on this PDB. For the CDP formula, two models were used, $CDP_1$ and $CDP_2$, which denote the 1-step and 2-step models, respectively.

As outlined in Section 4.1.1, the conditional distribution tables were built by generating one billion states (each is generated by applying 180 random moves to the goal state), computing all their neighbors, and incorporating their heuristic information into the matrix representing the one-step model. For the two-step model we also generated all the grandchildren and used their heuristic information.

In addition, in order to get reliable samples we added the following two procedures:

- While generating children and grandchildren for sampling we used the same pruning techniques based on operator ordering that were used in the main search (see the description in Section 2.1.1). That is, we did not use a sequence of operators that would not be generated by the main search. This is done by looking at the random walk that led to the initial state and using the last operator in the random walk as the basis for operator pruning.

- In order to get a reliable sample we need each entry in the table to be sufficiently sampled. Some entries in the table have very low frequency. For example, states with heuristic value of 0 are very rare even in a sample of a billion states causing our table for the 0 row to be generated by a very small sample. Therefore, we enriched such entries by artificially creating random states with heuristic value of 0. Other under-sampled entries were sampled in a similar way. One technique, for example, for creating (with high probability) a random state with a heuristic value of $x$, is to perform a random walk of length $x$ on a random state with heuristic value of 0.

### 6.1 Rubik's Cube with Consistent Heuristics

Table 6 compares KRE to $CDP_1$ and to $CDP_2$. The accuracy of the three prediction methods was compared while using regular lookups on the 6-edge PDB. Results in each row are averages over a set of 1000 random states. Each row presents the results of an IDA* iteration with different threshold ($d$), given in the first column. The second column (IDA*) presents the actual number of nodes expanded for each IDA* threshold. The next columns report the predictions and the accuracy ("Ratio") of each prediction defined as the ratio between the predicted number and the actual number of expanded nodes. As was reported in the KRE paper, the KRE formula was found to be very accurate for a *consistent* heuristic when averaged over a large set of random start states. The table shows that $CDP_1$ is reasonably accurate but systematically underestimates because the one-step model does not consider that a node's parent will not be included among its children. We elaborate on this below. $CDP_2$'s predictions are very accurate, slightly more accurate than KRE's.

| | | KRE | | $CDP_1$ | | $CDP_2$ | |
|---|---|---|---|---|---|---|---|
| d | IDA* | Prediction | Ratio | Prediction | Ratio | Prediction | Ratio |
| 8 | 277 | 257 | 0.93 | 235 | 0.85 | 257 | 0.93 |
| 9 | 3,624 | 3,431 | 0.95 | 3,151 | 0.87 | 3,446 | 0.95 |
| 10 | 47,546 | 45,801 | 0.96 | 41,599 | 0.87 | 45,985 | 0.97 |
| 11 | 626,792 | 611,385 | 0.98 | 546,808 | 0.87 | 613,332 | 0.98 |
| 12 | 8,298,262 | **8,161,064** | 0.98 | 7,188,863 | 0.87 | 8,180,676 | 0.99 |
| 13 | 110,087,215 | 108,937,712 | 0.99 | 94,711,234 | 0.86 | 109,133,021 | 0.99 |

Table 6: Rubik's Cube with a consistent heuristic.

## 6.2 Rubik's Cube with Start States Having Specific Heuristic Values

Table 2, presented above (Section 3.2.2), and the related discussion, show that KRE might not make accurate predictions when start states are restricted to have a specific heuristic value $h$. For the particular example shown (IDA* threshold 12) KRE will always predict a value of $8,161,064$, but the exact value depends on the specific set of start states used because the IDA* threshold of 12 is not sufficiently large for the number of nodes to be independent of the start states. Table 7 extends Table 2 to include the predictions of CDP. It shows that both versions of CDP substantially outperform KRE on any particular set of start states.

| | | KRE | | $CDP_1$ | | $CDP_2$ | |
|---|---|---|---|---|---|---|---|
| h | IDA* | Prediction | Ratio | Prediction | Ratio | Prediction | Ratio |
| 5 | 30,363,829 | 8,161,064 | 0.27 | 48,972,619 | 1.61 | 20,771,895 | 0.68 |
| 6 | 18,533,503 | 8,161,064 | 0.44 | 17,300,476 | 0.93 | 13,525,425 | 0.73 |
| 7 | 10,065,838 | 8,161,064 | 0.81 | 7,918,821 | 0.79 | 9,131,303 | 0.91 |
| 8 | 6,002,025 | 8,161,064 | 1.36 | 5,094,018 | 0.85 | 6,743,686 | 1.12 |
| 9 | 3,538,964 | 8,161,064 | 2.31 | 3,946,146 | 1.12 | 5,240,425 | 1.48 |

Table 7: Results for different start state heuristic values ($h$) for a regular PDB with an IDA* threshold of $d = 12$.

## 6.3 Rubik's Cube with Inconsistent Heuristics

The same experiments were repeated for inconsistent heuristics. The *dual* and *random-symmetry* lookups were performed on the 6-edge PDB instead of the regular lookup, thereby creating an inconsistent heuristic. As discussed in Section 3.2.1, KRE produces the same prediction for all heuristics (consistent and inconsistent) derived from a single PDB and overestimates for the inconsistent heuristics. Table 8 shows that $CDP_2$ is extremely accurate. Its prediction is always within 2% of the actual number of nodes expanded.

The 1-step model used by $CDP_1$ systematically underestimates the actual number of nodes expanded for regular and dual lookups (see the regular lookup in Table 6 and the dual lookup in Table 8). To understand why, consider what happens when the node $m$ in the right side of Figure 7 is expanded. It generates two children, node $n$ and (assuming operators have inverses as is the case in Rubik's Cube) a copy of its parent $R$ (shown as $m$'s left child in Figure 7). This child is 2 levels deeper than $R$ and therefore has an $f$-value that is 2 greater than $R$'s. With an IDA* threshold of 5, this child will not be a potential node and the 1-step model will conclude that $m$ will generate a potential child with a probability of 0.5, whereas in fact all of the children that remain after operator pruning are potential nodes.

| d | IDA* | KRE | | CDP₁ | | CDP₂ | |
|---|---|---|---|---|---|---|---|
| | | Prediction | Ratio | Prediction | Ratio | Prediction | Ratio |
| **Dual** | | | | | | | |
| 8 | 36 | 257 | 7.14 | 31 | 0.86 | 36 | 1.00 |
| 9 | 518 | 3,431 | 6.62 | 418 | 0.81 | 508 | 0.98 |
| 10 | 6,809 | 45,801 | 6.73 | 5,556 | 0.82 | 6,792 | 1.00 |
| 11 | 92,094 | 611,385 | 6.64 | 74,037 | 0.80 | 90,664 | 0.98 |
| 12 | 1,225,538 | 8,161,064 | 6.66 | 987,666 | 0.81 | 1,210,225 | 0.99 |
| 13 | 16,333,931 | 108,937,712 | 6.67 | 13,180,960 | 0.81 | 16,154,640 | 0.99 |
| **Random Symmetry** | | | | | | | |
| 8 | 26 | 257 | 9.88 | 26 | 1.00 | 26 | 1.00 |
| 9 | 346 | 3,431 | 9.92 | 353 | 1.02 | 346 | 1.00 |
| 10 | 4,608 | 45,801 | 9.94 | 4,718 | 1.02 | 4,601 | 1.00 |
| 11 | 61,617 | 611,385 | 9.92 | 62,990 | 1.02 | 61,174 | 0.99 |
| 12 | 823,003 | 8,161,064 | 9.92 | 840,849 | 1.02 | 815,444 | 0.99 |
| 13 | 10,907,276 | 108,937,712 | 9.99 | 11,224,108 | 1.03 | 10,878,227 | 1.00 |

Table 8: Rubik's Cube with dual, and random-symmetry (*inconsistent*) heuristics

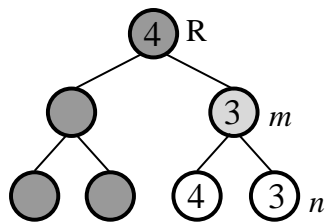

Figure 7: The 1-step model may underestimate

The reason the 1-step model does not underestimate the number of nodes expanded when random-symmetry lookups are done is because the child copy of $R$ is not constrained to have the same heuristic value as $R$ itself – different symmetries could be chosen for different occurrences $R$. The child's $f$-value has no correlation with the $f$-value of $R$ and the above explanation of why CDP₁ underestimates does not apply.

In fact, if different copies of a state have uncorrelated $h$-values the only effect of operator pruning that needs to be taken into account is that it reduces the number of children, and this can be done as well within a 1-step model when calculating the branching factor. There may be other advantages of using the wider context of the 2-step model but the results for the random-symmetry heuristic show that they are minor in this case.

## 7. Experimental Results - Sliding-Tile Puzzle

In the `KRE` experiments on the sliding-tile puzzle, three state types are used, based on whether the blank is in a corner, edge, or interior location. We used the same state types in our experiments and used exact recurrence equations for $N(s, v, d, t)$ in the type-dependent version of the `KRE` formula. The heuristic used was Manhattan Distance (MD). We experimented with the 2-step `CDP` that includes the type system in the recurrence equations. Results for the 1-step `CDP` are not included here because it performed poorly in early versions of these experiments.

For the 8-puzzle the conditional distribution $P(v, t|v_p, t_p, v_{gp}, t_{gp})$ needed by `CDP`$_2$ and the typed unconditional distribution $P(v, t)$ needed by the type-dependent `KRE` formula were computed by enumerating all the states in the 8-puzzle reachable from the goal.

For the 15-puzzle, it was not possible to do exhaustive enumeration of the entire state space so the conditional distributions were estimated by generating ten billion reachable states at random. This uniform random sample was used to estimate $P(v, t)$ for `KRE`, and each state in the sample was used as $gp$ in the sampling method described in Section 4.1.2 for $P(v, t|v_p, t_p, v_{gp}, t_{gp})$. For the latter, however, the basic sampling method had to be extended because even after processing ten billion $gp$ states some of the entries in the 6-dimensional matrix were missing or were not sampled sufficiently. To correct this, after we generate $gp$, its children, and its grandchildren and update the matrix accordingly, we check if the matrix already contains data for $gp$'s great-grandchildren. If it does not then we generate $gp$'s great-grandchildren and update the corresponding entries in the matrix. This continues as long as we encounter contexts that have never been seen before. This introduces a small statistical bias into the sample, but it guarantees that the sample contains the required data.

| | | | KRE | | CDP$_2$ | |
| $h$ | #States | IDA* | Prediction | Ratio | Prediction | Ratio |
|---|---|---|---|---|---|---|
| **8-puzzle depth 22** | | | | | | |
| 12 | 11,454 | 1,499 | 1,391 | 0.93 | 1,809 | 1.21 |
| 14 | 19,426 | 1,042 | 1,404 | 1.35 | 1,051 | 1.01 |
| 16 | 18,528 | 660 | 1,419 | 2.15 | 544 | 0.82 |
| 18 | 10,099 | 377 | 1,447 | 3.84 | 246 | 0.65 |
| 20 | 2,719 | 168 | 1,503 | 8.95 | 91 | 0.54 |
| **15-puzzle depth 52** | | | | | | |
| 34 | 1,331 | 77,028,888 | 420,858,250 | 5.46 | 172,845,559 | 2.24 |
| 36 | 2,330 | 38,206,986 | 424,113,561 | 11.10 | 64,247,275 | 1.68 |
| 38 | 2,999 | 16,226,330 | 428,883,700 | 26.43 | 21,505,426 | 1.33 |
| 40 | 3,028 | 6,310,724 | 433,096,514 | 68.63 | 6,477,903 | 1.03 |
| 42 | 2,454 | 2,137,488 | 438,475,079 | 205.14 | 1,749,231 | 0.82 |
| 44 | 1,507 | 620,322 | 444,543,678 | 716.63 | 409,341 | 0.66 |

Table 9: sliding-tile puzzles with a consistent heuristic (MD).

Prediction results for `KRE` and `CDP`$_2$ for the 8- and 15-puzzles are shown in Table 9 in the same format as above. For the 8-puzzle the predictions were made for an IDA* threshold of 22 and each row corresponds to the group of **all** 8-puzzle states with the same heuristic value $h$ (shown in the first column) for which IDA* would actually have used threshold 22. The second column gives the number of states in each group. Clearly, as shown in the "IDA*" column, for states with higher initial heuristic values IDA* expanded a smaller number of nodes. This trend is not reflected in the `KRE` predictions since `KRE` does not take $h$ into account. For `KRE` the only difference between the attributes of different rows is the different type distribution for the given group. Thus, the predicted number of expanded nodes of `KRE` is very similar for all rows (around 1,400). The `CDP` formula takes the heuristic value of the start state into account and was able to predict the number of expanded nodes much better than `KRE`. The bottom part of Table 9 show results for the 15-puzzle for an IDA* threshold of 52. Similar tendencies are observed.

## 7.1 Inconsistent Heuristics for the Sliding-tile Puzzle

Our next experiment is for an inconsistent heuristic on the 8-puzzle. We defined two PDBs, one based on the location of the blank and tiles 1–4, the other based on the location of the blank and tiles 5–8. To create an inconsistent heuristic, only one of the PDBs was consulted by a regular lookup. The choice of PDB was made systematically, not randomly, based on the position of the blank. Different occurrences of the same state were guaranteed to do the same lookup but neighboring states were guaranteed to consult different PDBs and this causes inconsistency. The results are presented in Table 10 for a variety of IDA* thresholds. For each threshold the "Num" column indicates how many start states were used. The results show that `CDP`'s predictions are reasonably accurate, and very much more accurate than `KRE`'s which overestimate by up to a factor of 26.

| d | Num | IDA* | KRE | | CDP$_2$ | |
|---|---|---|---|---|---|---|
| | | | Prediction | Ratio | Prediction | Ratio |
| 18 | 44,243 | 14.5 | 80.4 | 5.56 | 10.4 | 0.72 |
| 19 | 40,773 | 22.2 | 151.5 | 6.82 | 16.1 | 0.73 |
| 20 | 60,944 | 27.4 | 244.2 | 8.91 | 20.2 | 0.74 |
| 21 | 48,888 | 43.3 | 459.0 | 10.59 | 32.1 | 0.74 |
| 22 | 60,345 | 58.5 | 734.4 | 12.55 | 44.0 | 0.75 |
| 23 | 40,894 | 95.4 | 1,383.6 | 14.50 | 72.5 | 0.76 |
| 24 | 42,031 | 135.7 | 2,200.6 | 16.21 | 103.4 | 0.76 |
| 25 | 22,494 | 226.7 | 4,155.3 | 18.33 | 174.2 | 0.77 |
| 26 | 18,668 | 327.8 | 6,569.9 | 20.04 | 251.0 | 0.77 |
| 27 | 7,036 | 562.0 | 12,475.0 | 22.20 | 432.2 | 0.77 |
| 28 | 4,131 | 818.4 | 19,515.7 | 23.85 | 618.8 | 0.76 |
| 29 | 762 | 1,431.7 | 37,424.6 | 26.14 | 1,074.8 | 0.75 |

Table 10: Inconsistent heuristic for the 8-puzzle.

Similar experiments were conducted for the 15-puzzle. Here, the first PDB was based on the location of the blank and tiles 1–7, the other was based on the location of the blank and tiles 9–15. Table 11 shows the results for IDA* thresholds from 48 to 55 (recall that the median solution length for this puzzle is 52). The numbers shown are averages over 50,000 start states. The `CDP` predictions for the 15-puzzle are considerably worse than for the 8-puzzle, but the `KRE` predictions have degraded much more. The reason for the inaccuracy of these predictions was discussed in Section 4.5.2. Much more accurate predictions are produced if the context is extended to include the heuristic value of both pattern databases, not just the one that the search algorithm actually consults.

| d | IDA* | KRE | | CDP$_2$ | |
|---|------|-----------|-------|------------|-------|
|   |      | Prediction | Ratio | Prediction | Ratio |
| 48 | 231,939.6 | 311,462,527.1 | 1,342.9 | 71,550.2 | 0.308 |
| 49 | 388,201.1 | 664,920,142.2 | 1,712.8 | 149,257.5 | 0.384 |
| 50 | 644,350.1 | 1,413,202,357.9 | 2,193.2 | 313,132.4 | 0.486 |
| 51 | 1,062,597.5 | 3,014,405,997.5 | 2,836.8 | 663,004.4 | 0.624 |
| 52 | 1,746,025.1 | 6,404,191,951.4 | 3,667.9 | 1,402,898.2 | 0.803 |
| 53 | 2,773,611.6 | 13,639,455,787.3 | 4,917.6 | 2,985,321.1 | 1.076 |
| 54 | 4,539,767.0 | 29,035,096,650.9 | 6,395.7 | 6,361,011.5 | 1.401 |
| 55 | 7,546,286.9 | 61,899,533,064.7 | 8,202.6 | 13,627,941.8 | 1.806 |

Table 11: Inconsistent heuristic for the 15-puzzle.

## 8. Accurate Predictions for Single Start States

We have seen that `CDP` works well when the base cases of the recursive calculation of $N_i(s, d, v)$ are seeded by a large set of start states, no matter how their heuristic values are distributed. However, the actual number of expanded nodes for a specific **single** start state can deviate from the number predicted by `CDP`. The conditional distribution reflects the expected values over all nodes that share the same context, and the single start state of interest might behave differently than the "average" state that has the same context. Consider a Rubik's Cube state with a heuristic value of 8. `CDP`$_2$ predicts that IDA* will expand $6,743,686$ for such a state with IDA* threshold 12. Table 2 shows that on the average (over $1,000$ start states with a heuristic value of 8) $6,002,025$ states are expanded. Examining the results for the individual start states showed that the actual number of expanded nodes ranged between $2,398,072$ to $15,290,697$ nodes.

In order to predict the number of expanded nodes for a single start state we propose the following enhancement to `CDP`. Suppose that we want to predict the number of expanded nodes for IDA* threshold $d$ and start state $s$. First, we perform a small initial search from $s$ to depth $r$. We then use all the states at depth $r$ to seed the base cases of the `CDP` formula and compute the formula with IDA* threshold $d - r$. This will cause a larger set of nodes to be used in calculating $N_i(s, d, v)$, thereby improving the accuracy of `CDP`'s predictions.

### 8.1 Rubik's Cube, 6-edge PDB Heuristic

Table 12 shows results for four specific Rubik's Cube states with a heuristic value of 8 (of the regular 6-edge PDB lookup) when the IDA* threshold was set to 12. We chose the states with the least and greatest number of expanded nodes and two states around the median. The first column shows the actual number of nodes IDA* expands for each state.

The next columns show the number of expanded nodes predicted by our enhanced CDP$_2$ formula where the initial search was performed to depths ($r$) of 0, 2, 5 and 6. Clearly, these initial searches give much better predictions than the original CDP$_2$ (with $r = 0$), which predicts $6,743,686$ for all these states. With an initial search to depth 6, the predictions are very accurate.

| $h$ | IDA* | CDP$_2$(r=0) | CDP$_2$(r=2) | CDP$_2$(r=5) | CDP$_2$(r=6) |
|---|---|---|---|---|---|
| 8 | 2,398,072 | 6,743,686 | 4,854,485 | 3,047,836 | 2,696,532 |
| 8 | 4,826,154 | 6,743,686 | 7,072,952 | 5,495,475 | 5,184,453 |
| 8 | 9,892,376 | 6,743,686 | 8,555,170 | 9,611,325 | 9,763,455 |
| 8 | 15,290,697 | 6,743,686 | 9,432,008 | 13,384,290 | 14,482,001 |

Table 12: Single state ($d = 12$).

## 8.2 Rubik's Cube, 8-6-6 Heuristic

Section 3.2.3 presented KRE predictions for two start states, $s_6$, with a heuristic value of 6, and $s_{11}$, with a heuristic value of 11, for Rubik's Cube with the 8-6-6 heuristic. Here we repeat these experiments with CDP$_1$. Tables 13 and 14 show the results with an initial search of depth ($r$) 0 and 4. The tables show that CDP$_1$ was able to achieve substantially better predictions than KRE in most cases, and that an initial search to depth 4 usually improved CDP$_1$'s predictions.

| $d$ | IDA* | KRE | Ratio | CDP$_1$ (r=0) | Ratio | CDP$_1$(r=4) | Ratio |
|---|---|---|---|---|---|---|---|
| 10 | 53,262 | 1,510 | 0.03 | 32,207 | 0.60 | 69,770 | 1.31 |
| 11 | 422,256 | 20,169 | 0.05 | 246,158 | 0.58 | 690,556 | 1.64 |
| 12 | 3,413,547 | 269,229 | 0.08 | 1,979,417 | 0.58 | 5,422,001 | 1.59 |
| 13 | 29,114,115 | 3,593,800 | 0.12 | 16,690,055 | 0.57 | 42,650,077 | 1.46 |
| 14 | 259,577,913 | 47,971,732 | 0.18 | 149,319,061 | 0.58 | 345,370,148 | 1.33 |
| 15 | 2,451,954,240 | 640,349,193 | 0.26 | 1,435,177,445 | 0.59 | 2,934,134,125 | 1.20 |
| 16 | 24,484,797,237 | 8,547,681,506 | 0.35 | 14,925,206,678 | 0.61 | 26,380,507,927 | 1.08 |
| 17 | 258,031,139,364 | 114,098,463,567 | 0.44 | 167,181,670,892 | 0.65 | 254,622,231,216 | 0.99 |

Table 13: 8-6-6 PDB, single start state $s_6$

## 8.3 Experiments on the 8-Puzzle - Single Start States

We performed experiments with the enhanced CDP$_2$ formula on all the states of the 8-puzzle with the (consistent) MD heuristic. We use the term "trial" to refer to each pair of a single start state and a given IDA* threshold $d$. The trials included all possible values of $d$ and for each $d$ all start states for which IDA* would actually perform a search with IDA* threshold $d$. Predictions were made for each trial separately, and the relative error, *predicted/actual*, for the trial was calculated. The results are shown in Figure 8. There are four curves in the figure, for KRE, for CDP, and for the enhanced CDP with initial search depths ($r$) of 5 and 10. The $x$-axis is relative error. The $y$-axis is the percentage of trials for which the prediction had a relative error of $x$ or less. For example, the $y$-value of 20% for the KRE curve at $x = 0.5$ means that KRE underestimated by a factor of 2 or more on 20% of the trials. The rightmost point of the KRE plot ($x = 10$, $y = 94\%$) indicates that on 6% of the trials KRE's prediction was more than 10 times the actual number of nodes expanded. By contrast CDP has a much larger percentage of highly accurate predictions, with over 99% of its predictions within a factor of two of the actual number of nodes expanded. The figure clearly shows the advantage of using the enhanced CDP. With an initial search to a depth of 10, 90% of the trials had predictions within 10% of the correct number.

| $d$ | IDA* | KRE | Ratio | CDP$_1$ (r=0) | Ratio | CDP$_1$(r=4) | Ratio |
|---|---|---|---|---|---|---|---|
| 11 | 8,526 | 20,169 | 2.37 | 8,246 | 0.97 | 8,904 | 1.04 |
| 12 | 162,627 | 269,229 | 1.66 | 191,077 | 1.17 | 139,422 | 0.86 |
| 13 | 2,602,029 | 3,593,800 | 1.38 | 3,188,470 | 1.23 | 2,834,542 | 1.09 |
| 14 | 38,169,381 | 47,971,732 | 1.26 | 47,281,091 | 1.24 | 45,690,554 | 1.20 |
| 15 | 542,241,315 | 640,349,193 | 1.18 | 665,292,864 | 1.23 | 614,042,865 | 1.13 |
| 16 | 7,551,612,957 | 8,547,681,506 | 1.13 | 9,125,863,883 | 1.21 | 8,544,807,943 | 1.13 |
| 17 | 103,934,322,960 | 114,098,463,567 | 1.10 | 123,571,401,411 | 1.19 | 120,978,148,822 | 1.16 |

Table 14: 8-6-6 PDB, single start state $s_{11}$

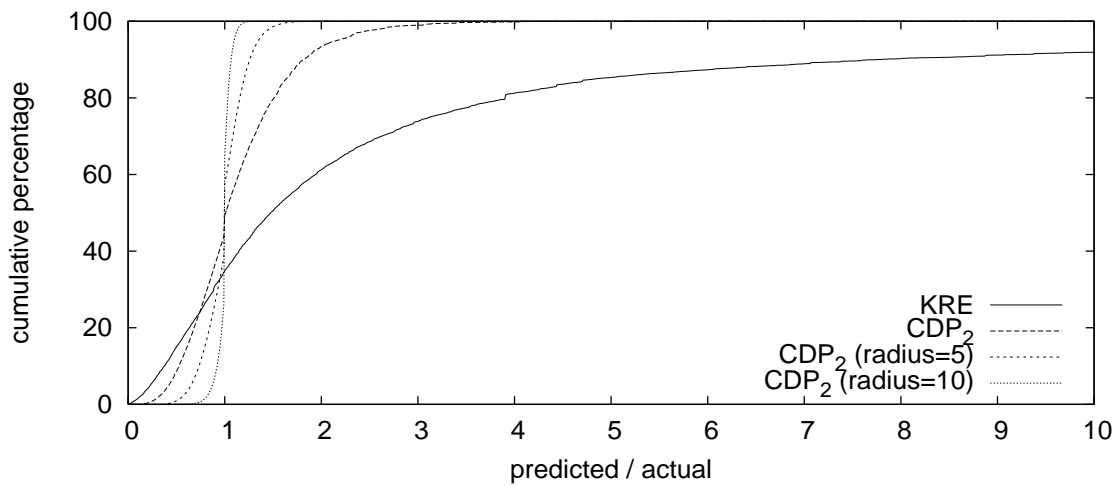

Figure 8: Relative error for the 8-puzzle

## 9. Performance Range for a Given Unconditional Distribution

The experiments in this paper that have used the 6-edge PDB for Rubik's Cube have illustrated the fact that the number of nodes IDA* expands given a PDB can vary tremendously depending on how the PDB is used (Zahavi et al., 2007). To see this clearly, the middle three columns of Table 15 show data that has already been seen in Tables 6 and 8, namely, the number of nodes IDA* expands when the 6-edge PDB is used in the regular manner, with dual lookups, and with random-symmetry lookups. IDA* expands ten times fewer nodes when the 6-edge PDB is consulted with random-symmetry lookups than when it is consulted in the normal way.

This raises the intriguing question of what range of performance can be achieved by varying the conditional distribution when the unconditional distribution is fixed.

| d | $\overline{\text{CDP}}$ | Regular | Dual | Random Symmetry | $\underline{\text{CDP}}$ |
|---|---|---|---|---|---|
| 8 | 257 | 277 | 36 | 26 | 16 |
| 9 | 3,431 | 3,624 | 518 | 346 | 210 |
| 10 | 45,801 | 47,546 | 6,809 | 4,608 | 2,813 |
| 11 | 611,385 | 626,792 | 92,094 | 61,617 | 37,553 |
| 12 | 8,161,064 | 8,298,262 | 1,225,538 | 823,003 | 501,273 |
| 13 | 108,937,712 | 110,087,215 | 16,333,931 | 10,907,276 | 6,691,215 |
| Correlation | | 0.591 | 0.359 | 0.187 | |

Table 15: Range of IDA* Performace for the 6-edge Rubik's Cube PDB

## 9.1 Upper Limit

The upper extreme, which results in the most nodes expanded, occurs when a consistent heuristic is used. This is because IDA* only expands potential nodes, so the maximum number of nodes is expanded when the conditional distribution is such that the parent of every potential node at level $i$ is a potential node at level $i-1$. An exact calculation of the number of potential nodes in the brute-force tree is therefore a theoretical upper bound on the number of nodes IDA* will expand for a given unconditional distribution. As we have already discussed, one way to estimate the number of potential nodes is to use the KRE formula. This estimate of the upper bound of the number of nodes that IDA* could expand is denoted as $\overline{\text{CDP}}$ in Table 15.

Alternatively, the number of potential nodes can be approximated with the CDP formula given the conditional distribution. Consider Equation 6. In the summation we consider all possible $v_p$ values in $[0, d-(i-1)]$ as only these nodes are potential nodes at level $i-1$. Thus only these nodes are expanded by IDA* at level $i-1$ and only these nodes can generate children at level $i$.[6] Now, let's substitute this with $v_p \in [0, h_{max}]$. We now consider *all* the nodes at level $i-1$, even the ones that are not potential nodes. Using this in the summation will calculate the number of *all* nodes with heuristic $v$ at level $i$ even ones that are not actually generated be IDA* (because their parents were not potential nodes, i.e. with $v_p > d - (i-1)$). This is shown in Equation 13.

$$\overline{N_i}(s,v) = \sum_{v_p=0}^{h_{max}} \overline{N_{i-1}}(s,v_p) \cdot b_{v_p} \cdot p(v|v_p) \tag{13}$$

---

6. Note that if the heuristic is consistent then only $v_p$ values in $\{v-1, v, v+1\}$ need to be considered in the summation because nodes with other values of $v_p$ (smaller than $v-1$ or larger than $v+1$) cannot generate children with a heuristic value of $v$.

Using this in the general prediction equation we get:

$$\overline{\text{CDP}} = \sum_{i=0}^{d} \sum_{v=0}^{d-i} \overline{N_i}(s, v) \tag{14}$$

This gives an alternative method to approximate the number of potential nodes. Both these methods approximate this upper bound. In practice, however, it is possible that the number of expanded nodes will slightly exceed this approximate bound due to noise and small errors in the sampling or the calculations.

## 9.2 Lower Limit

With consistent heuristics values of neighboring states are highly correlated. At the other extreme are cases where there is no correlation between heuristic values of neighboring nodes. That is, the heuristic value of a child node is statistically independent of the heuristic value of its parent. This means that regardless of the parent's heuristic value $v_p$, the heuristic values of the children are distributed according to the unconditional heuristic distribution, *i.e.*, $p(v|v_p) = p(v)$.

Our motivation for using this as an estimated lower bound on the number of nodes IDA* could expand for a given unconditional distribution is the empirical observation that the number of nodes IDA* expands decreases as the correlation between a parent's heuristic value and its children's heuristic values decreases.

This is illustrated in the last row of the three middle columns of Table 15, which shows the correlation between the heuristic values of neighboring states when different types of lookups are done in the 6-edge PDB. It was calculated using Pearson's correlation coefficient, defined over $n$ pairs of $x, y$ values according to the following equation

$$Correlation_{xy} = \frac{n \sum_{i=1}^{n} x_i y_i - \sum_{i=1}^{n} x_i \sum_{i=1}^{n} y_i}{\sqrt{n \sum_{i=1}^{n} x_i^2 - (\sum_{i=1}^{n} x_i)^2} \sqrt{n \sum_{i=1}^{n} y_i^2 - (\sum_{i=1}^{n} y_i)^2}} \tag{15}$$

In order to calculate the correlation, $60,000$ random pairs of $(x_i, y_i)$ neighboring states were generated. Their heuristic values were computed and used in Equation 15. The bottom row of Table 15 shows that the number of nodes expanded decreases as the correlation between neighboring heuristic values decreases. This leads us to suggest that the number of nodes expanded will reach a minimum when the correlation is zero.[7]

This estimated lower bound can be calculated using the `CDP` formula with $p(v|v_p) = p(v)$. We denote this by CDP. For the 1-step model this would be calculated using the following equations:

---

7. In theory, it is possible for a heuristic to have a *negative* correlation between the parent's heuristic value and its children's heuristic values, *i.e.*, parents with low heuristic values could tend to have children with large heuristic values and vice versa. We believe this is unlikely to occur in practice.

$$\underline{N_i}(s,d,v) = \sum_{v_p=0}^{d-(i-1)} \underline{N_{i-1}}(s,d,v_p) \cdot b_{v_p} \cdot p(v) \tag{16}$$

$$\underline{CDP} = \sum_{i=0}^{d} \sum_{v=0}^{d-i} \underline{N_i}(s,d,v) \tag{17}$$

As can be seen by comparing the rightmost two columns in Table 15, the random-symmetry use of the 6-edge PDB is within a factor of two of our estimated minimum possible number of nodes expanded with this PDB, which suggests that to substantially improve upon its performance one would have to use a different PDB.

Table 16 shows the estimated upper and lower bounds of IDA*'s performance, for a range of IDA* thresholds, for three different PDBs for Rubik's Cube. The bounds are calculated using 1,000 random start states. The table shows that, according to these estimates, inconsistent heuristics based on the 5-edge PDB can outperform consistent heuristics based on the 6-edge PDB but probably cannot outperform consistent heuristics based on the 7-edge PDB since the estimated lower bound of the 5-edge PDB is larger than the estimated upper bound of the 7-edge PDB.

| | 5-edge PDB | | 6-edge PDB | | 7-edge PDB | |
|---|---|---|---|---|---|---|
| d | $\overline{CDP}$ | $\underline{CDP}$ | $\overline{CDP}$ | $\underline{CDP}$ | $\overline{CDP}$ | $\underline{CDP}$ |
| 8 | 2,869 | 134 | 257 | 16 | 42 | 10 |
| 9 | 38,355 | 2,278 | 3,431 | 210 | 348 | 42 |
| 10 | 511,982 | 30,623 | 45,801 | 2,813 | 4,535 | 291 |
| 11 | 6,834,185 | 408,775 | 611,385 | 37,553 | 60,535 | 3,829 |
| 12 | 91,225,920 | 5,456,512 | 8,161,064 | 501,273 | 808,051 | 51,116 |
| 13 | 1,217,726,395 | 72,836,079 | 108,937,712 | 6,691,215 | 10,786,252 | 682,311 |

Table 16: Estimated Bounds on Performance for three Rubik's Cube PDBs.

## 10. Predicting the Performance of IDA* with BPMX

With an inconsistent heuristic, the heuristic value of a child can be much larger than that of the parent. If this happens in a state space with undirected edges, the child's heuristic value can be propagated back to the parent. If this causes the parent's $f$-value to exceed the IDA* threshold the entire search subtree rooted at the parent can be pruned without generating any of the remaining children. This propagation technique is called bidirectional pathmax (BPMX) (Felner et al., 2005; Zahavi et al., 2007). It was shown to be very effective in reducing the search effort by pruning subtrees that would otherwise be explored. We now show how to modify `CDP` to handle BPMX propagation. Since BPMX only applies to state spaces with undirected edges, the discussion in this section is limited to such spaces.

## 10.1 Bidirectional Pathmax (BPMX)

Traditional pathmax (Méro, 1984) propagates heuristic values from a parent to its children, and can be applied in any state space. Admissibility is preserved by subtracting the cost of the connecting edge from the heuristic value. The basic insight of bidirectional pathmax (BPMX) is that when edges are undirected heuristic values can propagate to all neighbors, which includes from a child node to its parent. This process can continue to any distance in any direction. BPMX is illustrated in Figure 9. The left side of the figure shows the inconsistent heuristic values for a node and its two children. Consider the left child with a heuristic value of 5. Since this value is admissible and all edges in this example have a cost of one, all its immediate neighbors are at least 4 moves away from the goal, their neighbors are at least 3 moves away, and so on. When the left child is generated, its heuristic value ($h = 5$) can propagate up to the parent and then down again to the right child. To preserve admissibility, each propagation along a path reduces $h$ by the cost of traversing the path. This results in $h = 4$ for the root and $h = 3$ for the right child. When using IDA*, this bidirectional propagation may cause many nodes to be pruned that would otherwise be expanded. For example, suppose the current IDA* threshold is 2. Without the propagation of $h$ from the left child, both the root node ($f = g + h = 0 + 2 = 2$) and the right child ($f = g + h = 1 + 1 = 2$) would be expanded. Using the above propagation, the left child will increase the parent's $h$ value to 4, resulting in search at this node being abandoned without even generating the right child.

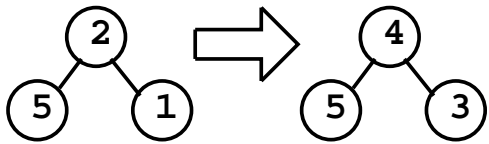

Figure 9: Propagation of values with inconsistent heuristics

## 10.2 `CDP` Overestimates When BPMX is Applied

When an inconsistent heuristic is being used and BPMX is applied, `CDP` will overestimate the number of expanded nodes because it will count the nodes in subtrees that BPMX prunes. In Section 4.2, we defined $N_i(s, d, v)$ to be the number of nodes that IDA* will generate at level $i$ with a heuristic value exactly equal to $v$ when $s$ is the start state and $d$ is the IDA* threshold. The formula given for estimating $N_i(s, d, v)$ (Equation 5) was:

$$\tilde{N}_i(s, d, v) = \sum_{v_p=0}^{d-(i-1)} \tilde{N}_{i-1}(s, d, v_p) \cdot b_{v_p} \cdot p(v|v_p)$$

In calculating $\tilde{N}_i(s, d, v)$ from $\tilde{N}_{i-1}(s, d, v_p)$ this formula assumes that when a node is expanded **all** its children are generated. This is why $\tilde{N}_{i-1}(s, d, v_p)$ is multiplied by the branching factor $b_{v_p}$. When BPMX is applied, a child may prune the parent before the rest of the children are generated. If this happens, the assumption that **all** the children of expanded nodes are generated would be wrong. For example, without BPMX, while expanding the root of the left tree in Figure 9 both children are generated and the child on the right is also expanded. Indeed CDP will count two nodes in this case. When BPMX is applied the root is expanded but the child on the right will not be generated (and therefore not expanded). Thus, CDP, which counts the two nodes, is overestimating the number of nodes expanded. In the following section we modify our equation to correct this.

### 10.3 New Formula for Estimating $N_i(s, d, v)$

Let $n$ be the node that is currently being expanded. Assume that $n$ has $b$ children and consider the order in which they are generated. We call this order the *generation order*. Note that when BPMX is applied, the probability that a child will be generated decreases as we move through the generation order. Children that appear late in the order will have a larger chance of not being generated since there are more previous children that might cause a BPMX cutoff. Let $p_{bx}(l)$ be the probability that the child in location $l$ in the order will be generated even if BPMX is applied. With this definition we can extend Equation 5 as follows:

$$\tilde{N}_i(s, d, v) = \sum_{v_p=0}^{d-(i-1)} \sum_{l=1}^{b_{v_p}} \{\tilde{N}_{i-1}(s, d, v_p) \cdot p_{bx}(l) \cdot p(v|v_p)\} \tag{18}$$

$\tilde{N}_i(s, d, v)$ is being calculated in a similar way to Equation 5, except for the way we count the total number of children IDA* generates via the nodes it expands at level $i-1$ with heuristic value equal to $v_p$. The idea here is to iterate over all the possible locations in the generation order and calculate the probability that a node in location $l$ will be generated. In practice, however, the actual context for $p_{bx}$ has other variables besides the location $l$. It also includes the IDA* threshold ($d$), the depth of the parent ($i-1$) and the heuristic value of the parent ($v_p$), We thus get our final formula:

$$\tilde{N}_i(s, d, v) = \sum_{v_p=0}^{d-(i-1)} \sum_{l=1}^{b_{v_p}} \{\tilde{N}_{i-1}(s, d, v_p) \cdot p_{bx}(l, d, i-1, v_p) \cdot p(v|v_p)\} \tag{19}$$

This is exactly equal to Equation 5 in the special case when $p_{bx}(l) = 1$ for all $l$, which happens when BPMX is not used or when it is used with a consistent heuristic.

### 10.4 Calculating $p_{bx}$

For simplicity, our model assumes that a heuristic value can only be propagated by BPMX one level up the tree. This means that a state can be pruned only from its immediate children and not by descendants at deeper levels. We make this assumption for another reason besides simplicity of description. Our experiments with Rubik's Cube and other domains showed that indeed almost all the pruning of BPMX was caused by a 1-level BPMX propagation. A generalized formula with deeper BPMX propagations can be similarly developed but it will include complicated recursive terms with very low practical value, at least for the state spaces and heuristics we have studied.

Assume that $c$ is a child of $n$ in location $l$ of the generation order. Child $c$ will be generated only if $n$ is not pruned by any of its $l-1$ children that appear before $c$ in the generation order. Assume that $n$ is at level $i$ and that the threshold is $d$. Since $n$ is expanded, $h(n) \leq d - i$. With BPMX $h(n)$ can be increased (and cause BPMX pruning) from a child $k$ if $h(k) > d - i + 1$. In this case, $h(k) - 1$ is larger than $d - i$, so when it is used instead of $h(n)$ IDA* will decide not to expand $n$ and no additional children will be generated. Therefore, in order for a child $c$ in location $l$ of the generation order to be generated, all its $l - 1$ predecessors in the generation order must have heuristics *less than or equal to* $d - i + 1$. Assuming that the heuristic value of the parent is $v$ the probability of this will be

$$p_{bx}(l, d, i, v) = \{ \sum_{h=0}^{d-i+1} p(h|v) \}^{l-1} \qquad (20)$$

We sum up the probability of each relevant heuristic value and raise the sum to the power of $l - 1$ since $l - 1$ children appear before $c$.

## 10.5 Experiments on Rubik's Cube with BPMX

We repeated the experiments on Rubik's Cube with the 6-edge PDB but with BPMX activated. Since BPMX affects only inconsistent heuristics, only the "Dual" and "Random Symmetry" heuristics were tested. Each heuristic was tested for IDA* thresholds 8 through 13. The results, averaged over the same set of 1,000 random states, are presented in Table 17. The "No BPMX" columns are repeated from Table 8. The additional columns show our results with BPMX. The column "IDA* + BPMX" presents the actual number of expanded nodes when using BPMX. BPMX reduces the number of nodes expanded by more than 30% for the Dual and by more than 25% reduction for the Random Symmetry, making the unmodified $\texttt{CDP}_2$'s predictions high by about the same amount. The "$\texttt{CDP}_2^{\texttt{bx}}$" column shows that the modifications introduced in this section greatly improve the accuracy.

| | No BPMX | | | With BPMX | | |
|---|---|---|---|---|---|---|
| d | IDA* | $\texttt{CDP}_2$ | Ratio | IDA* + BPMX | $\texttt{CDP}_2^{\texttt{bx}}$ | Ratio |
| **Dual** | | | | | | |
| 8 | 36 | 36 | 1.00 | 26 | 24 | 0.92 |
| 9 | 518 | 508 | 0.98 | 353 | 328 | 0.93 |
| 10 | 6,809 | 6,792 | 1.00 | 4,700 | 4,387 | 0.93 |
| 11 | 92,094 | 90,664 | 0.98 | 62,405 | 58,562 | 0.94 |
| 12 | 1,225,538 | 1,210,225 | 0.99 | 831,362 | 781,704 | 0.94 |
| 13 | 16,333,931 | 16,154,640 | 0.99 | 11,091,676 | 10,434,547 | 0.94 |
| **Random Symmetry** | | | | | | |
| 8 | 26 | 26 | 1.00 | 19 | 18 | 0.95 |
| 9 | 346 | 346 | 1.00 | 256 | 240 | 0.94 |
| 10 | 4,608 | 4,601 | 1.00 | 3,432 | 3,207 | 0.93 |
| 11 | 61,617 | 61,174 | 0.99 | 45,881 | 42,818 | 0.93 |
| 12 | 823,003 | 815,444 | 0.99 | 608,816 | 571,556 | 0.94 |
| 13 | 10,907,276 | 10,878,227 | 1.00 | 8,125,962 | 7,629,396 | 0.94 |

Table 17: BPMX on Rubik's Cube - Dual & Random Symmetry

## 11. Related Work

Previous work on predicting A* or IDA*'s performance from properties of a heuristic falls into two main camps. The first bases its analysis on the accuracy of the heuristic, while the second bases its analysis, as we have done, on the distribution of heuristic values. The next two subsections survey these approaches.

### 11.1 Analysis Based on a Heuristic's Accuracy

One common approach is to characterize a heuristic by focusing on the error in the heuristic value (deviation from the optimal cost). The first analysis in this line, focusing on the effect of errors on the performance of search algorithms, was done by Pohl (1970). Many other papers in this line have appeared since (Pohl, 1977; Gaschnig, 1979; Huyn, Dechter, & Pearl, 1980; Karp & Pearl, 1983; Pearl, 1984; Chenoweth & Davis, 1991; McDiarmid & Provan, 1991; Sen, Bagchi, & Zhang, 2004; Dinh, Russell, & Su, 2007; Helmert & Röger, 2008).

These works usually assume an abstract model space of a tree where every node has exactly $b$ children and aim to provide the asymptotic estimation for the number of expanded nodes. They mainly differ by the model assumptions (e.g. binary or non-binary trees) and for what case the results are derived (worst case or average case). Worst case analysis showed that there is a correlation between the heuristic errors and the search complexity. They found that if the relative error, $\frac{|h(n)-h^*(n)|}{h^*(n)}$, is constant, the search complexity will be exponential (in the length of solution path) but if the absolute error, $|h(n) - h^*(n)|$, is bounded by a constant the search complexity is linear (Pohl, 1977; Gaschnig, 1979). Three main assumptions used by Pohl (1977) are that the branching factor is assumed to be constant across inputs, that there is a single goal state and that there are no transpositions in the search space. When these assumptions do not hold, as is the case for many standard benchmark domains in planning, general search algorithms such as A* explore exponential number of states even under the assumption of an almost perfect heuristic (i.e., a heuristic whose error is bounded by a small additive constant) (Helmert & Röger, 2008).

Since it is difficult to guarantee precise bounds on the magnitude of errors produced by a given heuristic, a probabilistic characterization of these magnitudes was suggested (Huyn et al., 1980; Pearl, 1984). Heuristics are modeled as random variables (RVs), and the relative errors are assumed to be independent and identically distributed (IID model). In this model, attaining an average polynomial A* complexity was proved to be essentially equivalent to requiring that values of $h(n)$ be clustered near $h^*(n)$ where the allowed deviation is a logarithmic function of $h^*(n)$ itself.

Additional research in this line was conducted by Chenoweth and Davis (1991). Instead of using the IID model, they suggested using the "NC model", which places no constraints on the errors of $h$. With this model the heuristic is defined according to how the heuristic values grow with respect to the distance to the goal, and not according to the error. They predicted that A* complexity will be polynomial whenever the values of $h(n)$ are logarithmicaly clustered near $h^*(n) + \eta(h^*(n))$, where $\eta$ is an arbitrary, non-negative, and non-decreasing function. Heuristics whose values grow slower than the distance to the goal cause exponential complexity. Studies with the "NC model" showed that replacing a heuristic $h$ with $wh$ for some $w \geq 0$ can often change A* complexity from exponential to polynomial.

Most of these works focused on tree searches. By contrast, Sen et al. (2004) presented a general technique for extending the analysis of the average case performance of A* from search spaces that are trees to search spaces that are directed acyclic graphs. Their analytical results show that the expected complexity can change from exponential to polynomial as the heuristic estimates of nodes become more accurate and restrictions are placed on the cost matrix. Recent research in this line, analyzing the complexity of the A* algorithm was presented by Dinh et al. (2007). This research presented both worst and average case analysis for the performance of A* for approximately accurate heuristics[8] for search problems with multiple solutions. Bounds presented in that paper have been proved to be dependent on the heuristic accuracy and distribution of solutions.

## 11.2 Analysis Based on the Heuristic Distribution

As discussed at the outset of this paper, `KRE` suggested an alternative approach for calculating the time complexity of IDA* on multiple-goal spaces (Korf & Reid, 1998; Korf et al., 2001). Arguing that the heuristic accuracy is very difficult to obtain, they suggested deriving the analysis from the unconditional distribution of heuristic values, which is easy to determine at least approximately. They also came up with a method for deriving a closed-form formula for $N_i$, the number of nodes at level $i$ of the brute-force search tree. That method was later formalized (Edelkamp, 2001b). Unlike the work described in the previous subsection, which provides a "big-O" complexity analysis, `KRE`'s aim (and ours) is to exactly predict the number of nodes IDA* will expand.

`KRE` correctly point out that, when operators do not all have the same cost, $N_i$ must be defined as the number of nodes that can be reached by a path of cost $i$, as opposed to the number of nodes that are $i$ edges from the start state. The calculation of $N_i$ in this more general setting has been studied in detail by Ruml, in a slightly different context (Ruml, 2002). His solution involves using a conditional distribution for edge costs that bears a strong resemblance to our conditional distribution on heuristic values.

Based on the work of `KRE` and on the insight that for PDB heuristics there is a correlation between the size of the PDB and its heuristic value distribution, a new analysis limited to PDB heuristics has been done (Korf, 2007; Breyer & Korf, 2008). The prediction is achieved based on the branching factor of the problem and the size of the PDB without knowing the actual heuristic distribution. In order to derive the heuristic distribution from the size of the PDB it was assumed that the forward and backward branching factors of the abstract space are equal and that the abstract space has a negligible number of cycles. Since the second assumption is usually not realistic this model underestimates the number of expanded nodes.

The `KRE` formula was developed to predict the performance of the IDA* algorithm. The general approach can also be applied to A* as long as appropriate modifications are made to the computations of $N_i$ and $P(v)$ (Korf et al., 2001; Holte & Hernádvölgyi, 2004; Breyer & Korf, 2008). The challenge is accounting for the effect of A*'s pruning of the search tree when it generates a state that it has previously reached by a path of smaller or equal cost. This is particularly challenging when the heuristic is inconsistent, because in that case the first time A* generates a state it is not guaranteed to have reached it via a least-cost path, so the state will occur more than once in A*'s search tree. Indeed, in the worst case, for every state A* will enumerate all the paths to the state in decreasing order of cost, thereby generating exactly the same search tree as IDA* (Martelli, 1977). But in general, A*'s pruning will reduce $N_i$, especially for large $i$, in ways that may be hard to capture in a small set of recurrence equations. The heuristic distribution over A*'s entire search tree, taken to its maximum depth, is, for consistent heuristics, the overall distribution (Korf et al., 2001) since each state occurs exactly once in A*'s search tree (as just observed, this is not true for inconsistent heuristics). This does not imply that the overall distribution can be used to good effect on a level-by-level basis, but its use in the `KRE` formula did result in accurate predictions of A*'s performance on the 15-puzzle for two different consistent heuristics when used together with an exact calculation of $N_i$ for A*'s search tree (Breyer & Korf, 2008).

---

8. A heuristic is an $\epsilon$-approximation if $(1 - \epsilon)h^*(s) \leq h(s) \leq (1 + \epsilon)h^*(s)$ for all states in the search space.

## 12. Conclusions and Future Work

Historically, heuristics were characterized by their average. `KRE` introduced the idea of characterizing heuristics by their unconditional heuristic distribution and presented their formula to predict the number of nodes expanded on one iteration of IDA* based on the unconditional heuristic distribution. The work we have presented in this paper takes another step along this line. The conditional distribution we have introduced, and the prediction formula `CDP` based on it, advance our understanding of how properties of a heuristic affect the performance of IDA*.

Our `CDP` method advances `KRE` by improving its predictions at shallow depths, on a wider range of sets of start states, and for inconsistent heuristics. We have also shown how to use it to make an accurate prediction for a single start state and for an IDA* search that uses BPMX heuristic value propagation.

Of course, with the more sophisticated methods, more preprocessing is needed and special care must be taken when gathering the data in order to get a reliable sample. It is much easier to calculate the average of the heuristic than to calculate a 3-dimensional matrix. On the other hand, the latter approach better characterizes the heuristic and enables generating accurate predictions for a larger variety of circumstances.

Future work will address a number of issues. It is not yet clear what attributes make the best context for prediction, and how this is influenced by the choice of the heuristic and by the attributes of the specific domain. Larger contexts (more parameters) will probably provide better prediction at a cost of more pre-processing. This tradeoff needs to be further studied. Another direction will aim to extend this analysis approach to predict the performance of other search algorithms such as A*.

## 13. Acknowledgments

This research was supported by grant number 728/06 and 305/09 from the Israeli Science Foundation (ISF) to Ariel Felner. Robert Holte and Neil Burch gratefully acknowledge the ongoing support for this work from Canada's Natural Sciences and Engineering Research

Council (NSERC) and Alberta's Informatics Circle of Research Excellence (iCORE). The code for Rubik's Cube in this paper is based on the implementation of Richard E. Korf used in his seminal work on this domain(Korf, 1997). We thank the anonymous reviewer who encouraged us to widen our experimental results and to better explain the results of KRE and their relation to our results. His/her comments clearly improved the strength of this paper. Thanks also to Sandra Zilles for her careful checking of the details in Section 4.